\documentclass[journal]{IEEEtran}
\usepackage{amsmath,amsfonts}
\usepackage{algorithmic}
\usepackage{algorithm}
\usepackage{array}
\usepackage{textcomp}
\usepackage{stfloats}
\usepackage{url}
\usepackage{verbatim}
\usepackage{graphicx}
\usepackage{cite}
\usepackage{booktabs} % This package is used for professional-quality tables
\usepackage{multirow} % This package is used for tabular cells spanning multiple rows
\usepackage{subcaption} % This package is used for subfigures and subtables
\usepackage{xcolor} % This package is used for textcolor

\usepackage[hidelinks]{hyperref} % This package is used for clickable references

\begin{document}

\title{CLIP-PING: Boosting Lightweight Vision-Language Models with Proximus Intrinsic Neighbors Guidance}

\author{Chu Myaet Thwal, Ye Lin Tun, Minh N. H. Nguyen,~\IEEEmembership{Member,~IEEE}, Eui-Nam Huh,~\IEEEmembership{Member,~IEEE}, and\\Choong Seon Hong,~\IEEEmembership{Fellow,~IEEE}
\thanks{Chu Myaet Thwal, Ye Lin Tun, Eui-Nam Huh, and Choong Seon Hong are with the Department of Computer Science and Engineering, Kyung Hee University, Yongin-si, Gyeonggi-do, 17104, South Korea. (email: \{chumyaet, yelintun, johnhuh, cshong\}@khu.ac.kr). Corresponding author: CS Hong.}
\thanks{Minh N. H. Nguyen is with the Digital Science and Technology Institute, The
University of Danang---Vietnam-Korea University of Information and Communication Technology, Da Nang, 550000, Vietnam (email: nhnminh@vku.udn.vn).}
}

\maketitle

%% Abstract
\begin{abstract}\label{sec:abstract}
%Trend
Beyond the success of Contrastive Language-Image Pre-training (CLIP), recent trends mark a shift toward exploring the applicability of lightweight vision-language models for resource-constrained scenarios.
%Problem
These models often deliver suboptimal performance when relying solely on a single image-text contrastive learning objective, spotlighting the need for more effective training mechanisms that guarantee robust cross-modal feature alignment.
%Solution
In this work, we propose \mbox{CLIP-PING}: Contrastive Language-Image Pre-training with Proximus Intrinsic Neighbors Guidance, a novel yet simple and efficient training paradigm designed to boost the performance of lightweight vision-language models with minimal computational overhead and lower data demands.
%Claim
\mbox{CLIP-PING} bootstraps unimodal features extracted from arbitrary pre-trained encoders to obtain intrinsic guidance of proximus neighbor samples, i.e., nearest-neighbor (NN) and cross nearest-neighbor (XNN).
We find that extra contrastive supervision from these neighbors substantially boosts cross-modal alignment, enabling lightweight models to learn more generic features with rich semantic diversity.
%Results
Extensive experiments reveal that \mbox{CLIP-PING} notably surpasses its peers in zero-shot generalization and cross-modal retrieval tasks.
Specifically, a \textbf{5.5\%} gain on zero-shot ImageNet1K classification with \textbf{10.7\%} (I2T) and \textbf{5.7\%} (T2I) on Flickr30K retrieval, compared to the original CLIP when using ViT-XS image encoder trained on 3 million (image, text) pairs.
Moreover, \mbox{CLIP-PING} showcases a strong transferability under the linear evaluation protocol across several downstream tasks.
\end{abstract}

%-------------------------------------------------------------------------


\section{Introduction}\label{sec:introduction}

\IEEEPARstart{R}{ecent advances} in multi-modal contrastive representation learning~\cite{zhang2022contrastive, radford2021learning, bao2022vlmo, jia2021scaling, qi2020imagebert, cherti2023reproducible} have unlocked the potential of vision-language foundation models to effectively learn visual concepts from natural language supervision.
Leveraging the complementary strengths of visual and textual data, Contrastive Language-Image Pre-training (CLIP)~\cite{radford2021learning} swiftly gained notable attention for its impressive zero-shot generalization capability and excellent transferability to a wide range of downstream tasks.
Yet, despite these strides, the computational burden and data-hungry nature of CLIP pose significant challenges to replicate and build upon these groundbreaking results~\cite{li2023scaling, li2024inverse, li2022supervision, wang2023too, mu2022slip}, consequently imposing a serious barrier to its widespread application in resource-constrained scenarios.

\begin{figure}[t]
  \centering
   \includegraphics[width=0.9\linewidth]{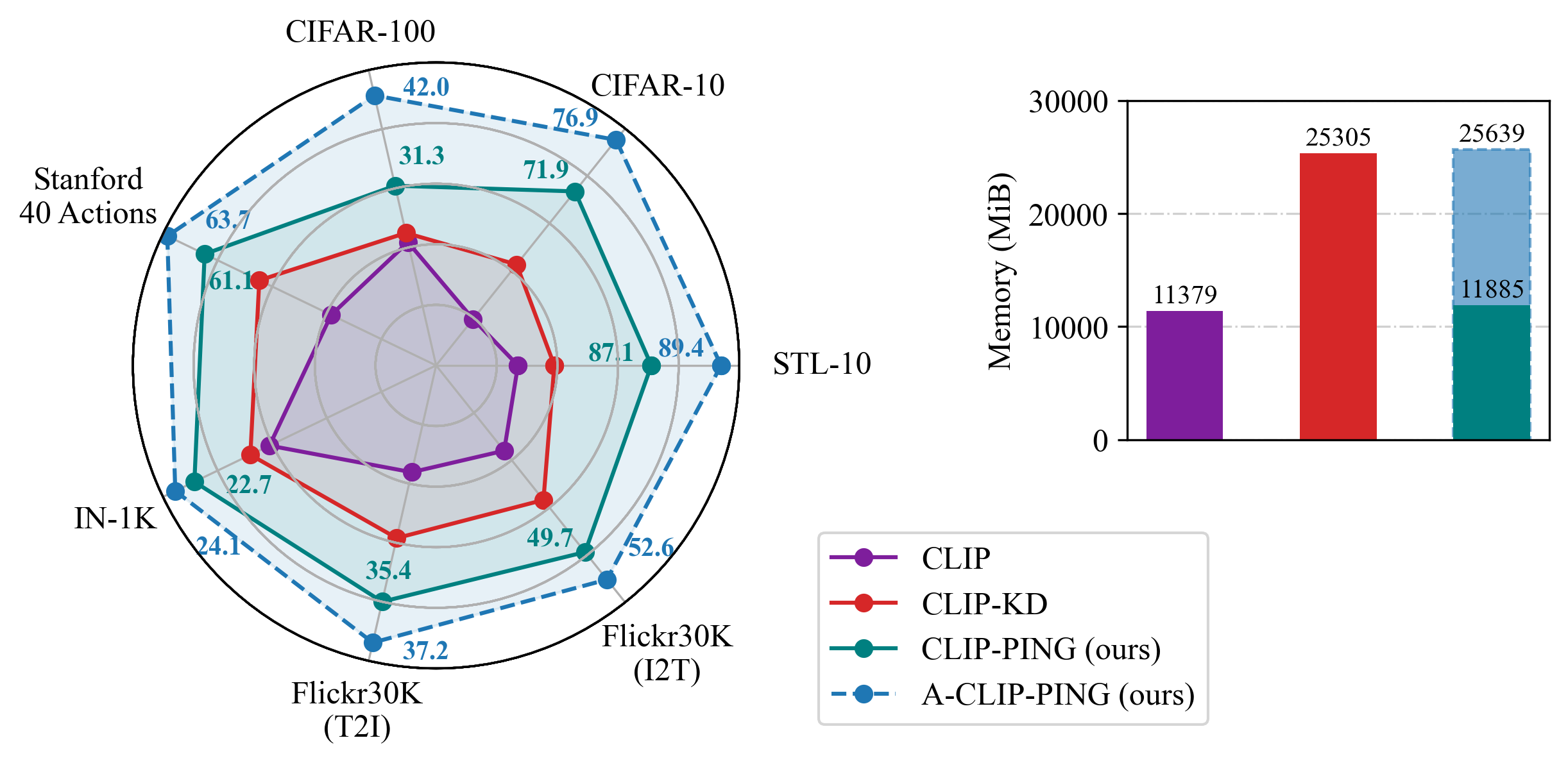}
   \caption{Comparison on zero-shot classification and retrieval performance using the ViT-XS~\cite{dosovitskiy2020image} image encoder, trained on COCO+CC3M~\cite{lin2014microsoft, sharma2018conceptual} dataset with 3M (image, text) pairs.}
   \label{fig:radar_chart}
   \vspace{-1em}
\end{figure}

So far, most studies focus on scaling up models and expanding data volumes to boost performance~\cite{wang2023image, jia2021scaling}; however, these approaches come at the cost of applicability.
This in turn demands substantial computational resources, leading to quadratic increases in training times across numerous high-powered devices, restricting accessibility to a limited group of researchers at large institutions and tech companies~\cite{li2022supervision, li2024inverse}.
As model sizes grow, hardware requirements also escalate, further narrowing the potential range of deployment for vision-language models.
Additionally, obtaining large quantities of high-quality paired data is often costly and challenging, particularly in domains where data privacy and confidentiality are crucial~\cite{karunarathna2024crucial}.
Evidently, this general trend toward large-scale language-image pre-training has become intractable for many practitioners.
On the other hand, standard small-scale language-image pre-training typically results in suboptimal performance, as it relies solely on a single image-text contrastive learning objective, overlooking additional supervision that can be further derived from the data itself.
These challenges trigger our efforts to develop more effective training mechanisms that strike a balance between model size, computational efficiency, and data requirements while preserving robust cross-modal alignment in resource-constrained settings.

\begin{figure}[t]
  \centering
   \includegraphics[width=0.8\linewidth]{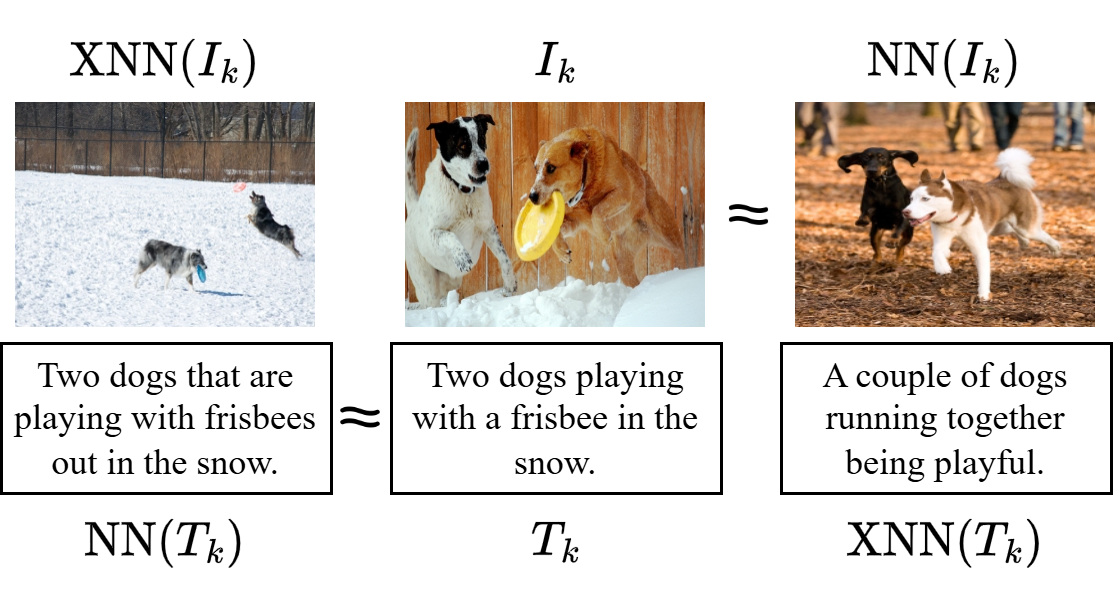}
   \caption{Example of nearest-neighbor~(NN) and cross nearest-neighbor~(XNN) samples for (image, text) pair, i.e., $(I_k,T_k)$, from the COCO~\cite{lin2014microsoft} dataset.}
   \label{fig:nn_xnn}
   \vspace{-1em}
\end{figure}

To this end, some studies have explored distillation strategies that enable lightweight vision-language models (i.e., students) to mimic the learning behavior of larger pre-trained models (i.e., teachers)~\cite{wu2023tinyclip, yang2024clipkd, vasu2024mobileclip, chen2024comkd, yang2024clip, liang2024module}.
Inspired by this concept, we introduce \textbf{\textit{\mbox{CLIP-PING}: Contrastive Language-Image Pre-training with Proximus Intrinsic Neighbors Guidance}}, aimed to boost the potential of lightweight models in resource-constrained scenarios.
In contrast to prior works on CLIP distillation~\cite{wu2023tinyclip, yang2024clipkd, chen2024comkd, yang2024clip}, \mbox{CLIP-PING} leverages features extracted from off-the-shelf pre-trained encoders, storing them frozen in auxiliary feature banks to provide intrinsic guidance from proximus neighbors without the need for explicit distillation.
To keep neighbor retrieval efficient, we maintain two representative support sets of frozen features --- one for each modality --- as manageable subsets of auxiliary feature banks.
This enables lightweight student encoders to capture rich knowledge of resource-intensive teacher encoders without extra computational burden or architectural constraints during training.

Beyond the typical image-text contrastive objective, \mbox{CLIP-PING} incorporates widespread supervision from semantically similar or neighboring samples across modalities.
Nearest-neighbor contrastive learning~\cite{han2020self, wu2018improving, dwibedi2021little, liu2023promoting} enables models to leverage supervision from similar samples within the data.
DeCLIP~\cite{li2022supervision} further explores nearest-neighbor retrieval across modalities for cross-supervision, based on the notion that one image may have multiple semantically related text descriptions.
Building on these insights, \mbox{CLIP-PING} draws on two primary sources of contrastive supervision from teacher encoders to boost the performance of student encoders.
First, it obtains \textbf{\textit{intra-modal supervision}} through nearest-neighbor (NN) samples of frozen features within each modality, encouraging feature alignment of similar images or text descriptions within the same feature space.
Second, it leverages cross nearest-neighbor (XNN) samples (also from the same modality) for \textbf{\textit{inter-modal supervision}}, encouraging the indirect alignment between semantically related pairs across modalities, by cross-referencing NN frozen features.
For instance, as shown in Fig.~\ref{fig:nn_xnn}, for an image $I_k$, its XNN is identified as the image associated with NN text description of its paired text $T_k$.
This dual-source supervision of \mbox{CLIP-PING} enables student encoders to learn more generic features with rich semantic diversity, while minimizing resource demands.

Our experiments show that \mbox{CLIP-PING} achieves superior performance over its counterparts in zero-shot classification and cross-modal retrieval tasks.
As illustrated in Fig.~\ref{fig:radar_chart}, using ViT-XS~\cite{dosovitskiy2020image} pre-trained on the combined COCO+CC3M~\cite{lin2014microsoft, sharma2018conceptual} dataset enables \mbox{CLIP-PING} to reach 22.7\% zero-shot top-1 accuracy on ImageNet1K~\cite{deng2009imagenet}, along with  49.7\% I2T and 35.4\% T2I retrieval R@1 on Flickr30K~\cite{young2014image}.~These results surpass the original CLIP~\cite{radford2021learning} by \textbf{5.5\%}, \textbf{10.7\%}, and \textbf{5.7\%}, respectively, without extra computational costs.
Notably, scaling up computational resources further enhances the performance.
In particular, \mbox{A-CLIP-PING}, which incorporates active teacher encoders for stronger guidance, delivers additional boosts by \textbf{1.4\%}, \textbf{2.9\%}, and \textbf{1.8\%}, with computational demands comparable to \mbox{CLIP-KD}~\cite{yang2024clipkd}.
\mbox{CLIP-PING} also demonstrates strong transferability in downstream tasks under the linear evaluation protocol. Our contributions are summarized as follows:

\begin{itemize}

    \item We propose \mbox{CLIP-PING}, a simple yet novel training mechanism, to boost performance of lightweight vision-language models in resource-constrained scenarios.
    
    \item We leverage extracted features from pre-trained unimodal encoders, providing intrinsic guidance through nearest-neighbor (NN) and cross nearest-neighbor (XNN). These features are frozen in auxiliary feature banks, enabling lightweight vision-language models to efficiently capture rich knowledge of pre-trained encoders without extra computational burden or architectural constraints.
    
    \item We explore \textbf{\textit{intra-modal supervision}} through NN samples to enhance feature alignment within the same modality. We also incorporate \textbf{\textit{inter-modal supervision}} through XNN samples, encouraging indirect alignment between semantically similar pairs across modalities.
    
    \item Our extensive experiments show efficacy and versatility of \mbox{CLIP-PING} across several benchmarks, highlighting its innovative and potential contributions to optimizing lightweight vision-language models for widespread deployment in resource-constrained scenarios.
    
\end{itemize}

%-------------------------------------------------------------------------

\section{Background and motivation}\label{sec:relatedwork}

In this section, we review the evolution of contrastive language-image pre-training, with a highlight on recent advances in efficient training strategies.
We also discuss the motivations that led to the development of \mbox{CLIP-PING}.

\begin{figure*}[t]
    \centering
    \includegraphics[width=0.75\linewidth]{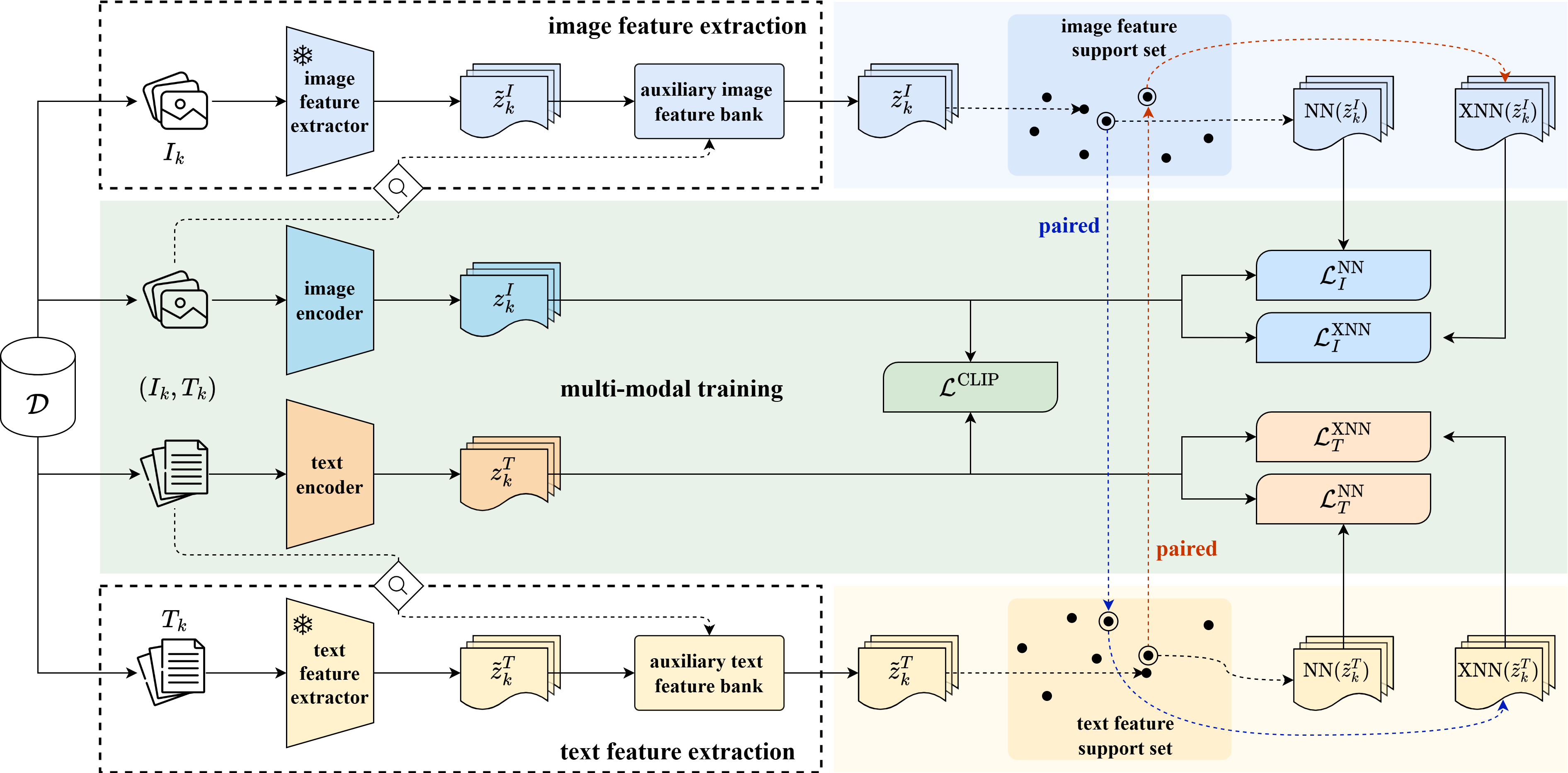}
    \caption{Overview of the \mbox{CLIP-PING} pipeline. Unimodal feature extraction is performed prior to the multi-modal training, with extracted features stored frozen in auxiliary feature banks. Each feature support set is a representative of the corresponding auxiliary feature bank.}
    \label{fig:clipping}
    \vspace{-1em}
\end{figure*}

%-------------------------------------------------------------------------

\subsection{Contrastive Language-Image Pre-training}

Learning transferable visual representations directly from natural language supervision has become increasingly predominant in computer vision~\cite{zhang2024vision, du2022survey, chen2023vlp}.
Early studies played on semantically dense captions to learn meaningful visual concepts over image-caption pairs~\cite{gomez2017self, gordo2017beyond, li2017learning, desai2021virtex, sariyildiz2020learning}.
Recent works leverage contrastive learning to align visual and textual representations in the shared latent space~\cite{jia2021scaling, qi2020imagebert, zhang2022contrastive, radford2021learning}.
Notably, Contrastive Language-Image Pre-training (CLIP)~\cite{radford2021learning} has gained significant traction due to its breakthrough in zero-shot multi-modal and unimodal visual tasks, utilizing the extensive WIT-400M private dataset.
While CLIP-like models benefit from large datasets containing millions or billions of (image, text) pairs on the Internet~\cite{pham2023combined, wang2021simvlm, yuan2021florence, cherti2023reproducible, zhai2022lit}, they rely heavily on large-scale training and typically demand substantial storage and computational resources, limiting applicability, especially in scenarios with limited data or resources.
Consequently, there is a growing need for effective training strategies that optimize the trade-off between performance and efficiency of CLIP-like models in resource-constrained settings.

%-------------------------------------------------------------------------

\subsection{Advances in efficient CLIP strategies}

While early iterations of CLIP-like models consistently yield better performance, they come with significant computational demands, as well as increased costs for large-scale data collection, storage, and processing~\cite{li2024inverse}.
To address these challenges, recent efforts have focused on making these models more accessible for widespread use in resource-constrained settings, like mobile and edge devices~\cite{wu2023tinyclip, vasu2024mobileclip}.
To this point, numerous studies have explored knowledge distillation as a key compression technique to enable efficient training of smaller models (i.e., students) under the supervision of larger pre-trained models (i.e., teachers)~\cite{fang2021compressing, wu2023tinyclip, yang2024clipkd, li2023distilling, wang2022multimodal}.
For instance, TinyCLIP~\cite{wu2023tinyclip} introduces cross-modal distillation through affinity mimicking and weight inheritance mechanisms, while CLIP-KD~\cite{yang2024clipkd} exploits several distillation strategies, including relational, feature-based, gradient-based, and contrastive paradigms.
Additionally, MobileCLIP~\cite{vasu2024mobileclip} introduces dataset reinforcement strategy to multi-modal setup, integrating knowledge from an ensemble of strong CLIP models and a pre-trained image captioning model to enhance learning efficiency.
These innovations mark a shift toward more efficient language-image pre-training, aimed at improving accessibility for real-world applications.

%-------------------------------------------------------------------------

\subsection{Research motivations}

Conversely, it is crucial for TinyCLIP~\cite{wu2023tinyclip} to share the same architectural-style between teacher and student models, limiting its flexibility for use with diverse architectures that might be better suited for specific tasks, such as those of specialized, lightweight student models.
Meanwhile, CLIP-KD~\cite{yang2024clipkd} involves balancing multiple complex distillation strategies across high-dimensional, multi-modal representations, which may lead to substantial computational and memory demands, posing challenges for implementation in low-resource settings.
Moreover, typical knowledge distillation frameworks require several forward passes through large pre-trained teacher models, which becomes infeasible when dealing with models containing billions or trillions of parameters.
These limitations raise several open research questions: 1) \textit{How can teacher and student models be effectively aligned without architectural constraints?} 2) \textit{Can contrastive losses be optimized for semantic alignment across modalities without explicit distillation?} 3) \textit{Is there an alternative to traditional knowledge distillation that can address these challenges, enabling more resource-efficient contrastive language-image pre-training?}
Thus, efficient training strategies with robust cross-modal alignment remain worth exploring.

%-------------------------------------------------------------------------

\section{\mbox{CLIP-PING} method}\label{sec:method}

In response to the above questions, we propose a novel yet simple and efficient training mechanism that enhances lightweight CLIP-like vision-language models using \textit{Proximus Intrinsic Neighbors Guidance} (PING), sourced from off-the-shelf, pre-trained encoders.
Unlike previous approaches, \mbox{CLIP-PING} uniquely combines \textit{intra-modal and inter-modal supervision} of frozen neighbor samples from pre-trained encoders via auxiliary feature banks.
This enables lightweight models to draw upon rich semantic knowledge of pre-trained unimodal encoders in a computationally efficient manner, without architectural constraints or explicit distillation.
In this section, we outline the \mbox{CLIP-PING} framework and describe its main components.
The overall \mbox{CLIP-PING} pipeline is structured in two stages: \textbf{unimodal feature extraction} and \textbf{multi-modal training}, as illustrated in Fig.~\ref{fig:clipping}.

%-------------------------------------------------------------------------

\subsection{Unimodal feature extraction}

\paragraph{Feature extractors}
Off-the-shelf encoders, pre-trained on large amounts of unimodal data, possess a rich semantic understanding of their respective modalities, providing a strong foundation for generating meaningful feature representations that can effectively guide lightweight models in multi-modal training.
As an initial step of \mbox{CLIP-PING}, we leverage these pre-trained encoders --- denoted as $\mathcal{F}_I^*$ and $\mathcal{F}_T^*$ --- as our image and text feature extractors (i.e., teachers).
Given a dataset of (image, text) pairs, $\mathcal{D}=\{(I_k,T_k)\}_{k=1}^{|\mathcal{D}|}$, we compute feature representations of teacher encoders for each modality prior to the multi-modal training stage.
Specifically, for each $I_k,T_k \in \mathcal{D}$, we obtain image features $\tilde{z}_k^I=\mathcal{F}_I^*(I_k)$ and text features $\tilde{z}_k^T=\mathcal{F}_T^*(T_k)$.
To minimize computational requirements, we process each modality separately on a single GPU, ensuring that only one large-scale unimodal encoder is loaded into memory at a time.
This preliminary feature extraction process enables efficient use of features generated by high-capacity teacher encoders, containing billions of parameters.

\paragraph{Auxiliary feature banks}
Once features are extracted from teacher encoders, they are stored frozen in reusable, auxiliary feature banks, i.e., $\mathcal{B}_I^*=\{\tilde{z}_k^I\}_{k=1}^{|\mathcal{D}|}$ for image and $\mathcal{B}_T^*=\{\tilde{z}_k^T\}_{k=1}^{|\mathcal{D}|}$ for text.
These feature banks allow frozen features to be accessed efficiently for neighbor retrieval, ensuring intrinsic guidance from the pre-trained teacher encoders during the student training.
This one-time unimodal feature extraction significantly streamlines the multi-modal training process, enabling computational efficiency, making \mbox{CLIP-PING} a potential resource-efficient solution for boosting the performance of lightweight vision-language models.

%-------------------------------------------------------------------------

\subsection{Multi-modal training}

\paragraph{Standard CLIP objective}
Following CLIP~\cite{radford2021learning}, we consider a dual-encoder architecture, comprising an image encoder $\mathcal{E}_I$ and a text encoder $\mathcal{E}_T$, to jointly learn meaningful feature representations across modalities.
For a batch of $N$ (image, text) pairs, i.e., $\{(I_k, T_k)\}_{k=1}^N$, image and text features, $z_k^I=\mathcal{E}_I(I_k)$ and $z_k^T=\mathcal{E}_T(T_k)$, are obtained through each encoder branch.
The objective is to learn the shared feature space, where corresponding images and texts are semantically aligned.
To this end, we employ the standard CLIP objective~\cite{radford2021learning} that ensures $N$ positive pairs to closely aligned while $N^2 - N$ irrelevant pairs remain apart.
Mathematically, this can be formalized as a symmetric function based on the InfoNCE~\cite{oord2018representation, zhang2022contrastive} loss, consisting of:
\begin{equation}
  \mathcal{L}_{I \rightarrow T}^{\text{CLIP}}=-\frac{1}{N}\sum_{k=1}^{N}\log\frac{\exp(\mathcal{S}(z_{k}^{I},z_{k}^{T})/\tau)}{\sum_{j=1}^{N}\exp(\mathcal{S}(z_{k}^{I},z_{j}^{T})/\tau)},
  \label{eq:clip_i2t}
\end{equation}
and
\begin{equation}
  \mathcal{L}_{T \rightarrow I}^{\text{CLIP}}=-\frac{1}{N}\sum_{k=1}^{N}\log\frac{\exp(\mathcal{S}(z_{k}^{T},z_{k}^{I})/\tau)}{\sum_{j=1}^{N}\exp(\mathcal{S}(z_{k}^{T},z_{j}^{I})/\tau)},
  \label{eq:clip_t2i}
\end{equation}
where $\mathcal{L}_{I \rightarrow T}^{\text{CLIP}}$ aligns image feature $z_k^I$ with its corresponding text feature $z_k^T$ by maximizing the similarity in-between, while $\mathcal{L}_{T \rightarrow I}^{\text{CLIP}}$ mirrors this.
The similarity $\mathcal{S}(\cdot, \cdot)$ is typically measured by the dot product between features, and scaled by a learnable temperature parameter $\tau$.
In summary, the overall image-text contrastive loss, $\mathcal{L}^{\text{CLIP}}$, is formulated as:
\begin{equation}
    \mathcal{L}^{\text{CLIP}}=\frac{1}{2}(\mathcal{L}_{I \rightarrow T}^{\text{CLIP}}+\mathcal{L}_{T \rightarrow I}^{\text{CLIP}}).
      \label{eq:clip}
\end{equation}

\begin{figure}[t]
    \centering
    \includegraphics[width=\linewidth]{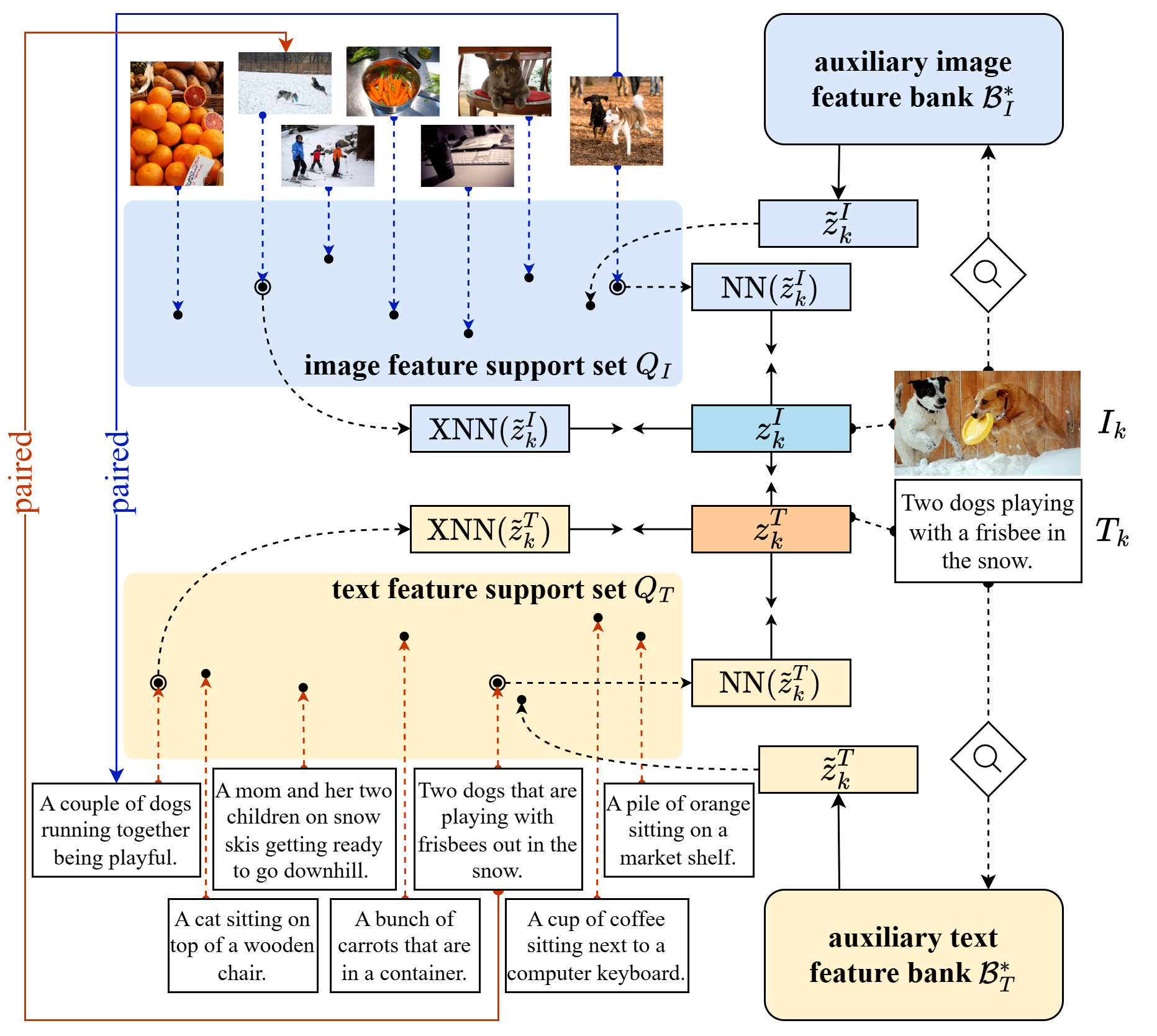}
    \caption{Example for nearest-neighbor (NN) and cross nearest-neighbor (XNN) retrieval process illustrated in a right-to-left order.}
    \label{fig:neighbors}
    \vspace{-1em}
\end{figure}

\paragraph{Intra-modal contrastive supervision through nearest-neighbors}
We leverage frozen features from each auxiliary feature bank (either image or text) to encourage each sample to be in close proximity with its semantically similar or \textit{nearest-neighbor (NN)} samples within the feature bank. 
To achieve this, we maintain two support sets, $\mathcal{Q}_I \subset \mathcal{B}_I^*$ and $\mathcal{Q}_T \subset \mathcal{B}_T^*$, capturing representative subsets of frozen features from auxiliary feature banks.
Specifically, image feature support set is $\mathcal{Q}_I = \{\tilde{z}_k^I\}_{k=1}^{|\mathcal{Q}_I|}$, and text feature support set is $\mathcal{Q}_T = \{\tilde{z}_k^T\}_{k=1}^{|\mathcal{Q}_T|}$.
The underlying concept of NN retrieval is shown in Fig.~\ref{fig:neighbors}, i.e., $\text{NN}(\tilde{z}):=\underset{q \in \mathcal{Q}}{\mathrm{argmin}}||\tilde{z}-q||_2$ for each modality.
Mathematically, intra-modal contrastive supervision through NN samples can be formalized as:
\begin{equation}
  \mathcal{L}_{\text{NN}_I \rightarrow I}=-\frac{1}{N}\sum_{k=1}^{N}\log\frac{\exp(\mathcal{S}(\text{NN}(\tilde{z}_k^I),z_{k}^{I})/\tau)}{\sum_{j=1}^{N}\exp(\mathcal{S}(\text{NN}(\tilde{z}_k^I),z_{j}^{I})/\tau)},
  \label{eq:ping_inn2i}
\end{equation}
and
\begin{equation}
  \mathcal{L}_{\text{NN}_T \rightarrow T}=-\frac{1}{N}\sum_{k=1}^{N}\log\frac{\exp(\mathcal{S}(\text{NN}(\tilde{z}_k^T),z_{k}^{T})/\tau)}{\sum_{j=1}^{N}\exp(\mathcal{S}(\text{NN}(\tilde{z}_k^T),z_{j}^{T})/\tau)},
  \label{eq:ping_tnn2t}
\end{equation}
where both losses are symmetric.
Thus, supervision from image NN is defined by:
\begin{equation}
    \mathcal{L}_I^{\text{NN}}= \frac{1}{2}(\mathcal{L}_{\text{NN}_I \rightarrow I} + \mathcal{L}_{I \rightarrow \text{NN}_I}).
    \label{eq:ping_inn}
\end{equation}
Likewise, supervision from text NN is defined by: 
\begin{equation}
    \mathcal{L}_T^{\text{NN}}= \frac{1}{2}(\mathcal{L}_{\text{NN}_T \rightarrow T} + \mathcal{L}_{T \rightarrow \text{NN}_T}).
    \label{eq:ping_tnn}
\end{equation}
The overall objective for intra-modal contrastive supervision through frozen NN samples, $\mathcal{L}_\text{NN}^{\text{PING}}$, is formulated as:
\begin{equation}
    \mathcal{L}_{\text{NN}}^{\text{PING}}=\mathcal{L}_I^{\text{NN}}+\mathcal{L}_T^{\text{NN}}.
      \label{eq:ping_nn}
\end{equation}
By incorporating intra-modal contrastive supervision in this way, \mbox{CLIP-PING} enables student encoders to capture rich semantic features of teacher encoders, which improves the robustness and strengthens alignment within each modality.

\paragraph{Inter-modal contrastive supervision through cross nearest-neighbors}
We explore \textit{cross nearest-neighbor (XNN)} samples within each modality, by cross-referencing frozen NN features to obtain more diverse supervisory signals from the dataset.
As illustrated in Fig.~\ref{fig:neighbors}, for a frozen image feature $\tilde{z}_k^I$, its XNN is identified as the image sample associated with NN of frozen text feature $\tilde{z}_k^T$, i.e., $\text{XNN}(\tilde{z}_k^I):=\tilde{z}_i^I \in \mathcal{Q}_I$, where $\tilde{z}_i^T=\text{NN}(\tilde{z}_k^T)$ for \mbox{$i$-th} pair $(\tilde{z}_i^I, \tilde{z}_i^T)$.
Likewise, for a frozen text feature $\tilde{z}_k^T$, $\text{XNN}(\tilde{z}_k^T):=\tilde{z}_i^T \in \mathcal{Q}_T$, where $\tilde{z}_i^I=\text{NN}(\tilde{z}_k^I)$ for $i$-th pair $(\tilde{z}_i^I, \tilde{z}_i^T)$.
Mathematically, inter-modal contrastive supervision through XNN samples can be formalized as:
\begin{equation}
  \mathcal{L}_{\text{XNN}_I \rightarrow I}=-\frac{1}{N}\sum_{k=1}^{N}\log\frac{\exp(\mathcal{S}(\text{XNN}(\tilde{z}_k^I),z_{k}^{I})/\tau)}{\sum_{j=1}^{N}\exp(\mathcal{S}(\text{XNN}(\tilde{z}_k^I),z_{j}^{I})/\tau)},
  \label{eq:ping_ixnn2i}
\end{equation}
and
\begin{equation}
  \mathcal{L}_{\text{XNN}_T \rightarrow T}=-\frac{1}{N}\sum_{k=1}^{N}\log\frac{\exp(\mathcal{S}(\text{XNN}(\tilde{z}_k^T),z_{k}^{T})/\tau)}{\sum_{j=1}^{N}\exp(\mathcal{S}(\text{XNN}(\tilde{z}_k^T),z_{j}^{T})/\tau)},
  \label{eq:ping_txnn2t}
\end{equation}
where both losses are symmetric. Thus, supervision from image XNN is defined by:
\begin{equation}
  \mathcal{L}_I^{\text{XNN}}= \frac{1}{2}(\mathcal{L}_{\text{XNN}_I \rightarrow I} + \mathcal{L}_{I \rightarrow \text{XNN}_I}).
  \label{eq:ping_ixnn}
\end{equation}
Likewise, supervision from text XNN is defined by:
\begin{equation}
  \mathcal{L}_T^{\text{XNN}}= \frac{1}{2}(\mathcal{L}_{\text{XNN}_T \rightarrow T} + \mathcal{L}_{T \rightarrow \text{XNN}_T}).
  \label{eq:ping_txnn}
\end{equation}
The overall objective for inter-modal contrastive supervision through frozen XNN samples, $\mathcal{L}_\text{XNN}^{\text{PING}}$, is formulated as:
\begin{equation}
    \mathcal{L}_{\text{XNN}}^{\text{PING}}=\mathcal{L}_I^{\text{XNN}}+\mathcal{L}_T^{\text{XNN}}.
      \label{eq:ping_xnn}
\end{equation}
By incorporating inter-modal contrastive supervision in this way, \mbox{CLIP-PING} encourages indirect alignment for semantically similar pairs of frozen features, enabling student encoders to capture richer latent features across modalities.

\paragraph{Supervision with PING objective}
Together, intra-modal and inter-modal supervision enable the learning of more generic features with enriched semantic diversity.
The overall PING objective can be summarized as:
\begin{equation}
    \mathcal{L}^{\text{PING}}=(1-\alpha) \cdot \mathcal{L}_{\text{NN}}^{\text{PING}}+\alpha \cdot \mathcal{L}_{\text{XNN}}^{\text{PING}}.
      \label{eq:ping}
\end{equation}
Here, $\alpha$ is a tunable hyperparameter to control the weight, balancing between intra-modal and inter-modal guidance.

\paragraph{Final \mbox{CLIP-PING} objective}
By integrating CLIP objective with PING supervision, we leverage the strengths of cross-modal contrastive alignment with rich, diverse supervision introduced by intrinsic neighbors (i.e., NN and XNN).
Thus, the final \mbox{CLIP-PING} objective, $\mathcal{L}^{\text{CLIP-PING}}$, is:
\begin{equation}
    \mathcal{L}^{\text{CLIP-PING}}=(1-\lambda) \cdot \mathcal{L}^{\text{CLIP}}+\lambda \cdot \mathcal{L}^{\text{PING}},
      \label{eq:final_loss}
\end{equation}
where $\lambda$ is a tunable hyperparameter that controls the contribution of $\mathcal{L}^{\text{PING}}$ relative to the standard CLIP loss, $\mathcal{L}^{\text{CLIP}}$.
This combined \mbox{CLIP-PING} objective effectively boosts lightweight models by leveraging knowledge from pre-trained unimodal encoders, while significantly reducing computational requirements, thus making \mbox{CLIP-PING} ideal for multi-modal training in resource-constrained settings. We provide the detailed procedure of \mbox{CLIP-PING} in the Supplementary Algorithm~1.

%-------------------------------------------------------------------------

\section{Experiments}\label{sec:experiments}

In this section, we provide implementation details and present extensive evaluation results that demonstrate the effectiveness of \mbox{CLIP-PING} across several benchmarks.

%-------------------------------------------------------------------------

\subsection{Implementation details}

We implement our experiments using PyTorch~\cite{paszke2019pytorch} and the Timm~\cite{rw2019timm} library.
Experiments using the dataset of 600K (image, text) pairs are run on a single NVIDIA RTX A6000 GPU with 48GB memory, while experiments using the dataset of 3M (image, text) pairs are conducted on a single NVIDIA A100 GPU with 40GB memory.

\paragraph{Training datasets}
We use the COCO~\cite{lin2014microsoft} dataset, which contains 600K (image, text) pairs, i.e., 118K images, each with 5 text descriptions.
We also explore another training set, where we combine COCO~\cite{lin2014microsoft} with the Conceptual Captions 3M (CC3M)~\cite{sharma2018conceptual}.
It is worth noting that, due to download issues, we only obtained 2.3M pairs from CC3M, bringing the total size of the combined COCO+CC3M~\cite{lin2014microsoft, sharma2018conceptual} dataset to about 3M (image, text) pairs.
As our primary focus is on resource-constrained scenarios, we intentionally avoid using larger Internet-scale datasets, commonly employed in several recent works, to prioritize computational efficiency and reduce storage demands.

\paragraph{Architectures}
We consider MobileBERT\textsubscript{TINY}~\cite{sun2020mobilebert} (with MobileBertTokenizer and a maximum context length of 55) as our text encoder and explore three variations of image encoders, each representing different architectural paradigms: ViT-XS~\cite{dosovitskiy2020image} (a smaller transformer-based model), ConvNeXt-Pico~\cite{liu2022convnet} (a compact convolutional network), and MNv4-Hybrid-M~\cite{qin2024mobilenetv4} (a hybrid convolutional transformer variant of MobileNetv4 model).
Each encoder is paired with a projection head, a two-layer MLP with GELU non-linearity for the first layer.
Specifications of all lightweight encoders are listed in Table~\ref{tab:encoders}.
Unless otherwise stated, we use ViT-XS + MobileBERT\textsubscript{TINY} as the default pair for ablations and analysis.

\paragraph{Baselines}
We compare \mbox{CLIP-PING} against the original CLIP~\cite{radford2021learning} and CLIP-KD~\cite{yang2024clipkd}, which combines feature distillation~(FD), interactive contrastive learning~(ICL), and contrastive relational distillation~(CRD).
For experiments with COCO~\cite{lin2014microsoft} dataset, we include additional comparisons with traditional CLIP distillation~(CLIP-D), and an efficient CLIP distillation using auxiliary feature banks~(CLIP-F).

\begin{table}[b]
\centering
\caption{Specifications of image and text encoders.}
\label{tab:encoders}
\resizebox{\linewidth}{!}{
\begin{tabular}{l|c|c|c|c|c}
\toprule
\multicolumn{1}{c|}{Image Encoder}              & Type              & \#Params              & \begin{tabular}[c]{@{}c@{}}Text Encoder\\ Transformer\end{tabular}                & \#Params              & \begin{tabular}[c]{@{}c@{}}Total\\ \#Params\end{tabular} \\ \midrule

ViT-XS~\cite{dosovitskiy2020image}              & ViT               & 8.3M                  & \multirow{4}{*}{\begin{tabular}[c]{@{}c@{}}MobileBERT\textsubscript{TINY} \\ \cite{sun2020mobilebert}\end{tabular}}        & \multirow{4}{*}{14.2M} & 22.5M                                                     \\ \cmidrule{1-3} \cmidrule{6-6} 
ConvNeXt-Pico~\cite{liu2022convnet}             & CNN               & 8.9M                  &                                                                                   &                        & 23.1M                                                      \\ \cmidrule{1-3} \cmidrule{6-6} 
MNv4-Hybrid-M~\cite{qin2024mobilenetv4}         & Hybrid           & 11.7M                  &                                                                                   &                        & 25.9M                                                       \\ \bottomrule
\end{tabular}
}
\end{table}

\paragraph{Unimodal feature extractors}
By default, we use \mbox{ResNet-v2-50} \cite{kolesnikov2020big, he2016identity} for image features and BERT-Base~\cite{kenton2019bert} (with BertTokenizer and a maximum context length of 55) for text, both initialized with pre-trained weights available on HuggingFace (i.e., timm/resnetv2\_50x1\_bit.goog\_in21k\_ft\_in1k and google-bert/bert-base-uncased).
Table~\ref{tab:feature_extractors} presents the specifications of pre-trained unimodal feature extractors used in our experiments, along with memory usage (MiB), feature bank size (GB), and total feature extraction time (hours) for each training dataset.
Each process was conducted with a batch size of 2048, on a single NVIDIA RTX A6000 GPU with 48GB memory.
Feature representations are extracted without any data augmentation.
The extracted features are stored in individual \texttt{pickle} files for efficient access during the training of \mbox{CLIP-F} and \mbox{CLIP-PING}.
During training, when there is a dimension difference between frozen features and those of lightweight encoders, a single-layer linear adapter is applied to the frozen features to match their dimensions.

\begin{table}[t]
\centering
\caption{Specifications of unimodal feature extractors.}
\label{tab:feature_extractors}
\resizebox{\linewidth}{!}{
\begin{tabular}{cccc|c|cc|cc}
\toprule
\multicolumn{4}{c|}{Feature Extractor}                                                                                                                              & \multirow{2}{*}{\begin{tabular}[c]{@{}c@{}}Memory\\ (MiB)\end{tabular}} & \multicolumn{2}{c|}{\begin{tabular}[c]{@{}c@{}}COCO~\cite{lin2014microsoft}\\ (600K)\end{tabular}}                                                      & \multicolumn{2}{c}{\begin{tabular}[c]{@{}c@{}}COCO+CC3M\\ \cite{lin2014microsoft, sharma2018conceptual}~(3M)\end{tabular}}                                                    \\ \cmidrule{1-4} \cmidrule{6-9} 
\multicolumn{1}{c|}{Image Encoder} & \multicolumn{1}{c|}{Type} & \multicolumn{1}{c|}{\#Params}                & \begin{tabular}[c]{@{}c@{}}Feat.\\ dim\end{tabular} &                                                                         & \multicolumn{1}{c|}{\begin{tabular}[c]{@{}c@{}}Size\\ (GB)\end{tabular}} & \begin{tabular}[c]{@{}c@{}}Time\\ (hours)\end{tabular} & \multicolumn{1}{c|}{\begin{tabular}[c]{@{}c@{}}Size\\ (GB)\end{tabular}} & \begin{tabular}[c]{@{}c@{}}Time\\ (hours)\end{tabular} \\ \midrule
\multicolumn{1}{c|}{ResNet-v2-50~\cite{kolesnikov2020big, he2016identity}}  & \multicolumn{1}{c|}{CNN}  & \multicolumn{1}{c|}{23.5M}                   & 2048                                                & 43734                                                                   & \multicolumn{1}{c|}{4.6}                                                 & 0.12                                                  & \multicolumn{1}{c|}{22.1}                                                & 0.56                                                 \\
\multicolumn{1}{c|}{ViT-B/16~\cite{dosovitskiy2020image}}      & \multicolumn{1}{c|}{ViT}  & \multicolumn{1}{c|}{85.8M}                   & 768                                                 & 21952                                                                   & \multicolumn{1}{c|}{1.7}                                                 & 0.38                                                 & \multicolumn{1}{c|}{8.5}                                                 & 1.85                                                \\ \midrule
\multicolumn{2}{c|}{Text Encoder: Transformer}                 & \multicolumn{1}{c|}{\multirow{2}{*}{109.5M}} & \multirow{2}{*}{768}                                & \multirow{2}{*}{7644}                                                   & \multicolumn{1}{c|}{\multirow{2}{*}{1.7}}                                & \multirow{2}{*}{0.13}                                 & \multicolumn{1}{c|}{\multirow{2}{*}{8.5}}                                & \multirow{2}{*}{0.60}                                \\
\multicolumn{2}{c|}{BERT-Base~\cite{kenton2019bert}}                         & \multicolumn{1}{c|}{}                        &                                                     &                                                                         & \multicolumn{1}{c|}{}                                                    &                                                      & \multicolumn{1}{c|}{}                                                    &                                                      \\ \bottomrule
\end{tabular}
}
\vspace{-1em}
\end{table}

\paragraph{Training details}
All lightweight models are trained from scratch for 35 epochs, applying a cosine learning rate scheduler with a linear warm-up over the first 5 epochs.
Training batch size is 1024 and the AdamW~\cite{loshchilov2018decoupled} optimizer with a weight decay of \mbox{1e-5} is used.
Initial learning rates of image and text encoders are \mbox{3e-3} and \mbox{1e-3}, respectively.
Input images are resized to $224 \times 224$, with \texttt{RandomResizedCrop} as the sole data augmentation used during training.
We use gradient checkpointing~\cite{griewank2000algorithm, chen2016training} and automatic mixed precision~\cite{micikevicius2018mixed} to improve memory efficiency and accelerate training.
Learnable temperature $\tau$ is initialized at 0.07, and the projection dimension is set to 256.
By default, we set the supervision loss weight values as $\alpha=0.25$ and $\lambda=0.6$.
Our support set is implemented as a first-in-first-out (FIFO) queue, with a queue size of $|Q|=32768$.
% More details are provided in the supplementary material.

%-------------------------------------------------------------------------

\subsection{Evaluation details}

\begin{table}[t]
\centering
\caption{Cross-modal retrieval performance for models trained on COCO~\cite{lin2014microsoft} (600K) dataset. The best results are marked in \textbf{bold}. Memory usage is recorded on a single NVIDIA RTX A6000 GPU.}
\label{tab:600K_retrieval}
\resizebox{\linewidth}{!}{
\begin{tabular}{lccccc}
\toprule
\multicolumn{1}{c|}{\multirow{2}{*}{Method}} & \multicolumn{1}{c|}{Memory} & \multicolumn{2}{c|}{COCO~\cite{lin2014microsoft}}        & \multicolumn{2}{c}{Flickr30K~\cite{young2014image}} \\
\multicolumn{1}{c|}{}                        & \multicolumn{1}{c|}{(MiB) $\downarrow$}                        & I2T@1 & \multicolumn{1}{c|}{T2I@1} & I2T@1         & T2I@1         \\ \midrule
\multicolumn{6}{c}{Model: ViT-XS~\cite{dosovitskiy2020image} + MobileBERT\textsubscript{TINY}~\cite{sun2020mobilebert}}                                                                                                          \\ \midrule
\multicolumn{1}{l|}{CLIP}                    & \multicolumn{1}{c|}{11074}                   & 20.4  & \multicolumn{1}{c|}{14.3}  & 19.1          & 14.7          \\
\multicolumn{1}{l|}{CLIP-D}          & \multicolumn{1}{c|}{25032}                   & 23.5  & \multicolumn{1}{c|}{16.0}  & 26.4          & 18.2          \\
\multicolumn{1}{l|}{CLIP-F}     & \multicolumn{1}{c|}{11192}                   & 21.2  & \multicolumn{1}{c|}{14.4}  & 23.9          & 16.5          \\
\multicolumn{1}{l|}{CLIP-KD}                 & \multicolumn{1}{c|}{25036}                   & 21.4  & \multicolumn{1}{c|}{15.6}  & 22.4          & 16.1          \\
\multicolumn{1}{l|}{CLIP-PING (ours)}        & \multicolumn{1}{c|}{11580}                   & \textbf{24.7}  & \multicolumn{1}{c|}{\textbf{18.4}}  & \textbf{28.1}          & \textbf{20.2}          \\
\multicolumn{1}{l|}{A-CLIP-PING (ours)}      & \multicolumn{1}{c|}{25370}                   & \textbf{27.6}  & \multicolumn{1}{c|}{\textbf{20.8}}  & \textbf{30.3}          & \textbf{22.3}          \\ \midrule
\multicolumn{6}{c}{Model: ConvNeXt-Pico~\cite{liu2022convnet} + MobileBERT\textsubscript{TINY}~\cite{sun2020mobilebert}}                                                                                                       \\ \midrule
\multicolumn{1}{l|}{CLIP}                    & \multicolumn{1}{c|}{16126}                   & 21.3  & \multicolumn{1}{c|}{15.9}  & 21.5          & 16.4          \\
\multicolumn{1}{l|}{CLIP-D}          & \multicolumn{1}{c|}{28456}                   & 26.3  & \multicolumn{1}{c|}{18.8}  & 28.1          & 20.1          \\
\multicolumn{1}{l|}{CLIP-F}     & \multicolumn{1}{c|}{16130}                   & 24.2  & \multicolumn{1}{c|}{17.1}  & 25.2          & 17.6          \\
\multicolumn{1}{l|}{CLIP-KD}                 & \multicolumn{1}{c|}{28460}                   & 24.7  & \multicolumn{1}{c|}{18.7}  & 24.6          & 18.7          \\
\multicolumn{1}{l|}{CLIP-PING (ours)}        & \multicolumn{1}{c|}{16874}                   & \textbf{27.3}  & \multicolumn{1}{c|}{\textbf{20.3}}  & \textbf{29.6}          & \textbf{20.8}          \\
\multicolumn{1}{l|}{A-CLIP-PING (ours)}      & \multicolumn{1}{c|}{28792}                   & \textbf{30.0}  & \multicolumn{1}{c|}{\textbf{22.2}}  & \textbf{31.9}          & \textbf{23.8}          \\ \midrule
\multicolumn{6}{c}{Model: MNv4-Hybrid-M~\cite{qin2024mobilenetv4} + MobileBERT\textsubscript{TINY}~\cite{sun2020mobilebert}}                                                                                           \\ \midrule
\multicolumn{1}{l|}{CLIP}                    & \multicolumn{1}{c|}{16696}                   & 21.7  & \multicolumn{1}{c|}{15.9}  & 21.1          & 16.9          \\
\multicolumn{1}{l|}{CLIP-D}          & \multicolumn{1}{c|}{31772}                   & 27.0  & \multicolumn{1}{c|}{19.8}  & 25.3          & 20.2          \\
\multicolumn{1}{l|}{CLIP-F}     & \multicolumn{1}{c|}{16696}                   & 24.1  & \multicolumn{1}{c|}{18.0}  & 22.7          & 17.3          \\
\multicolumn{1}{l|}{CLIP-KD}                 & \multicolumn{1}{c|}{31780}                   & 26.4  & \multicolumn{1}{c|}{19.9}  & 24.2          & 18.3          \\
\multicolumn{1}{l|}{CLIP-PING (ours)}        & \multicolumn{1}{c|}{17050}                   & \textbf{27.8}  & \multicolumn{1}{c|}{\textbf{21.1}}  & \textbf{29.9}          & \textbf{21.5}          \\
\multicolumn{1}{l|}{A-CLIP-PING (ours)}      & \multicolumn{1}{c|}{32110}                   & \textbf{31.9}  & \multicolumn{1}{c|}{\textbf{24.1}}  & \textbf{32.8}          & \textbf{24.9}          \\ \bottomrule
\end{tabular}
}
\vspace{-1em}
\end{table}

\paragraph{Downstream tasks, datasets, and metrics}
Models are evaluated across several downstream tasks, including cross-modal retrieval, zero-shot classification, and linear evaluation to assess the efficacy of \mbox{CLIP-PING}.
For cross-modal retrieval, we use the CC3M~\cite{sharma2018conceptual} validation set, i.e., 13K (image, text) pairs, and the COCO~\cite{lin2014microsoft} validation set, i.e., 5K images, each with 5 text descriptions.
Additionally, we use the Flickr30K~\cite{young2014image} test set, i.e., 1K images, each with 5 text descriptions, for zero-shot cross-modal retrieval.
Retrieval performance is measured with Recall@K metrics, i.e., R@1 for image-to-text (I2T) and text-to-image (T2I).
For zero-shot classification, evaluations are conducted on STL-10 (STL)~\cite{coates2011analysis}, CIFAR-10 (C10), CIFAR-100 (C100)~\cite{krizhevsky2009learning}, Stanford 40 Actions (SA-40)~\cite{yao2011human}, and ImageNet1K (IN-1K)~\cite{deng2009imagenet}, while we further demonstrate \mbox{CLIP-PING}'s robustness on additional 4 datasets, which are out of ImageNet distribution, such as ImageNet-V2 (IN-V2)~\cite{recht2019imagenet}, ImageNet-Rendition (IN-R)~\cite{hendrycks2021many}, ImageNet-O (IN-O)~\cite{hendrycks2021natural}, and ImageNet-Sketch (IN-S)~\cite{wang2019learning}.
We follow prompt engineering of the original CLIP paper~\cite{radford2021learning}, with specific prompt templates \texttt{"a photo of a person \{label\}"} and \texttt{"a photo of people \{label\}"} for the Stanford 40 Actions (SA-40)~\cite{yao2011human} dataset.
Linear evaluation is performed on 12 widely used downstream datasets, such as Oxford-IIIT Pets (Pets)~\cite{parkhi2012cats}, Caltech-101 (Cal.)~\cite{fei2004learning}, Oxford 102 Flowers (Flow.)~\cite{nilsback2008automated}, FGVC-Aircraft (FGVC)~\cite{maji2013fine}, Food-101 (Food)~\cite{bossard2014food}, Describable Textures (DTD)~\cite{cimpoi2014describing}, SUN397 (SUN)~\cite{xiao2010sun}, Stanford Cars (Cars)~\cite{krause2013collecting, krause20133d}, STL-10 (STL)~\cite{coates2011analysis}, CIFAR-10 (C10), CIFAR-100 (C100)~\cite{krizhevsky2009learning}, and ImageNet1K (IN-1K)~\cite{deng2009imagenet}. 
Top-1 accuracy is used as the primary metric for classification tasks, and we report the average accuracy across all datasets as ``AVG".

\paragraph{Cross-modal retrieval}
Tables~\ref{tab:600K_retrieval} and \ref{tab:3M_retrieval} show performance comparison on cross-modal retrieval task, highlighting \mbox{CLIP-PING}'s consistent superiority across all three lightweight vision-language models.
Table~\ref{tab:600K_retrieval} summarizes results for models trained on COCO~\cite{lin2014microsoft}, and Table~\ref{tab:3M_retrieval} provides results for the combined COCO+CC3M~\cite{lin2014microsoft,sharma2018conceptual} dataset.
Notably, \mbox{CLIP-PING} demonstrates robustness across various architectures.
The computational overhead of \mbox{CLIP-PING} is minimal compared to explicit distillation methods like CLIP-D and CLIP-KD~\cite{yang2024clipkd}, while remaining as efficient as the original CLIP~\cite{yang2024clip}.
Essentially, \mbox{CLIP-PING} achieves competitive performance without the resource-intensive processes required by distillation-based methods.
This efficiency likely stems from the way \mbox{CLIP-PING} leverages frozen pre-trained features, which reduces the need for repeated calculations typically associated with training.
We also observed that increasing computational resources further enhances the performance.
Specifically, we find that Active \mbox{CLIP-PING} (\mbox{A-CLIP-PING}) with active teacher encoders for stronger guidance during training, replacing the feature extraction stage, delivers additional performance boosts, for instance, \textbf{2.9\%} and \textbf{1.8\%} improvements on zero-shot Flickr30K~\cite{young2014image} retrieval for ViT-XS~\cite{dosovitskiy2020image} image encoder trained on COCO+CC3M~\cite{lin2014microsoft, sharma2018conceptual} dataset with computational demands remain comparable to CLIP-KD~\cite{yang2024clipkd}.

\begin{table}[t]
\centering
\caption{Cross-modal retrieval performance for models trained on COCO+CC3M~\cite{lin2014microsoft, sharma2018conceptual} (3M) dataset. The best results are marked in \textbf{bold}. Memory usage is recorded on a single NVIDIA A100 GPU.}
\label{tab:3M_retrieval}
\resizebox{\linewidth}{!}{
\begin{tabular}{lccccccc}
\toprule
\multicolumn{1}{c|}{\multirow{2}{*}{Method}} & \multicolumn{1}{c|}{Memory} & \multicolumn{2}{c|}{CC3M~\cite{sharma2018conceptual}}          & \multicolumn{2}{c|}{COCO~\cite{lin2014microsoft}}        & \multicolumn{2}{c}{Flickr30K~\cite{young2014image}} \\
\multicolumn{1}{c|}{}                       & \multicolumn{1}{c|}{(MiB) $\downarrow$}  & I2T@1 & \multicolumn{1}{c|}{T2I@1} & I2T@1 & \multicolumn{1}{c|}{T2I@1} & I2T@1         & T2I@1         \\ \midrule
\multicolumn{8}{c}{Model:  ViT-XS~\cite{dosovitskiy2020image} + MobileBERT\textsubscript{TINY}~\cite{sun2020mobilebert}}                                                                                                                                 \\ \midrule
\multicolumn{1}{l|}{CLIP}                    & \multicolumn{1}{c|}{11379}  & 23.6  & \multicolumn{1}{c|}{23.9}  & 32.6  & \multicolumn{1}{c|}{22.6}  & 39.0          & 29.7          \\
\multicolumn{1}{l|}{CLIP-KD}                 & \multicolumn{1}{c|}{25305}  & 26.3  & \multicolumn{1}{c|}{26.1}  & \textbf{35.1}  & \multicolumn{1}{c|}{24.3}  & 44.2          & 32.6          \\
\multicolumn{1}{l|}{CLIP-PING (ours)}        & \multicolumn{1}{c|}{11885}  & \textbf{26.4}  & \multicolumn{1}{c|}{\textbf{26.8}}  & 35.0  & \multicolumn{1}{c|}{\textbf{25.5}}  & \textbf{49.7}          & \textbf{35.4}          \\
\multicolumn{1}{l|}{A-CLIP-PING (ours)}      & \multicolumn{1}{c|}{25639}  & \textbf{27.9}  & \multicolumn{1}{c|}{\textbf{28.4}}  & \textbf{37.9}  & \multicolumn{1}{c|}{\textbf{27.0}}  & \textbf{52.6}          & \textbf{37.2}          \\ \midrule
\multicolumn{8}{c}{Model: ConvNeXt-Pico~\cite{liu2022convnet} + MobileBERT\textsubscript{TINY}~\cite{sun2020mobilebert}}                                                                                                                           \\ \midrule
\multicolumn{1}{l|}{CLIP}                    & \multicolumn{1}{c|}{16299}  & 23.1  & \multicolumn{1}{c|}{23.8}  & 35.2  & \multicolumn{1}{c|}{24.8}  & 45.1          & 33.1          \\
\multicolumn{1}{l|}{CLIP-KD}                 & \multicolumn{1}{c|}{28633}  & 28.0  & \multicolumn{1}{c|}{28.3}  & \textbf{39.8}  & \multicolumn{1}{c|}{\textbf{28.9}}  & 52.1          & \textbf{38.6}          \\
\multicolumn{1}{l|}{CLIP-PING (ours)}        & \multicolumn{1}{c|}{18615}  & \textbf{28.6}  & \multicolumn{1}{c|}{\textbf{28.9}}  & 37.8  & \multicolumn{1}{c|}{27.0}  & \textbf{52.7}          & 38.4          \\
\multicolumn{1}{l|}{A-CLIP-PING (ours)}      & \multicolumn{1}{c|}{28965}  & \textbf{29.6}  & \multicolumn{1}{c|}{\textbf{29.4}}  & \textbf{40.0}  & \multicolumn{1}{c|}{\textbf{29.3}}  & \textbf{54.4}          & \textbf{40.6}          \\ \midrule
\multicolumn{8}{c}{Model: MNv4-Hybrid-M~\cite{qin2024mobilenetv4} + MobileBERT\textsubscript{TINY}~\cite{sun2020mobilebert}}                                                                                                                           \\ \midrule
\multicolumn{1}{l|}{CLIP}                    & \multicolumn{1}{c|}{15747}  & 23.9  & \multicolumn{1}{c|}{24.3}  & 33.7  & \multicolumn{1}{c|}{24.2}  & 42.0          & 32.8          \\
\multicolumn{1}{l|}{CLIP-KD}                 & \multicolumn{1}{c|}{30831}  & 27.5  & \multicolumn{1}{c|}{27.6}  & 38.6  & \multicolumn{1}{c|}{27.1}  & 49.8          & 36.1          \\
\multicolumn{1}{l|}{CLIP-PING (ours)}        & \multicolumn{1}{c|}{16101}  & \textbf{28.1}  & \multicolumn{1}{c|}{\textbf{28.3}}  & \textbf{39.4}  & \multicolumn{1}{c|}{\textbf{28.6}}  & \textbf{52.0}          & \textbf{40.2}          \\
\multicolumn{1}{l|}{A-CLIP-PING (ours)}      & \multicolumn{1}{c|}{31161}  & \textbf{30.5}  & \multicolumn{1}{c|}{\textbf{30.2}}  & \textbf{41.5}  & \multicolumn{1}{c|}{\textbf{31.3}}  & \textbf{54.7}          & \textbf{41.6}          \\ \bottomrule
\end{tabular}
}
\vspace{-1em}
\end{table}

\paragraph{Zero-shot classification}
As shown in Table~\ref{tab:zs}, \mbox{CLIP-PING} consistently outperforms competing methods in zero-shot image classification across all datasets.
Specifically, it achieves average performance improvements of \textbf{7.5\%} and \textbf{4.7\%} over CLIP~\cite{radford2021learning} and CLIP-KD~\cite{yang2024clipkd}, respectively, with the ViT-XS~\cite{dosovitskiy2020image} image encoder trained on COCO+CC3M~\cite{lin2014microsoft, sharma2018conceptual} dataset.
The \mbox{A-CLIP-PING} variant provides an additional \textbf{4.4\%} performance boost.
Table~\ref{tab:robustness} demonstrates zero-shot robustness evaluation performance of ViT-XS~\cite{dosovitskiy2020image} image encoder trained on COCO+CC3M~\cite{lin2014microsoft, sharma2018conceptual} dataset, further showcasing \mbox{CLIP-PING}'s efficacy and superiority over other baselines.

\paragraph{Linear evaluation}
We train a linear classifier for 30 epochs with a batch size of 512.
The Adam~\cite{kingma2014adam} optimizer is used, with a cosine learning rate scheduler and an initial learning rate of 1e-2.
As shown in Table~\ref{tab:linear}, \mbox{CLIP-PING} outperforms other methods by a notable gap, illustrating average improvements of \textbf{9.1\%} and \textbf{4.8\%} over CLIP~\cite{radford2021learning} and CLIP-KD~\cite{yang2024clipkd}, respectively, when evaluating the ViT-XS~\cite{dosovitskiy2020image} image encoder trained on COCO+CC3M~\cite{lin2014microsoft, sharma2018conceptual} dataset.
Moreover, our \mbox{A-CLIP-PING} variant delivers an additional \textbf{2.8\%} performance boost.

\subsection{Ablations and analyses}
In this section, we conduct thorough analyses and ablation experiments to investigate the effectiveness of \mbox{CLIP-PING}.

\begin{figure}[b]
  \centering
  \begin{subfigure}[t]{0.23\textwidth}
  \centering
  \includegraphics[width=\textwidth]{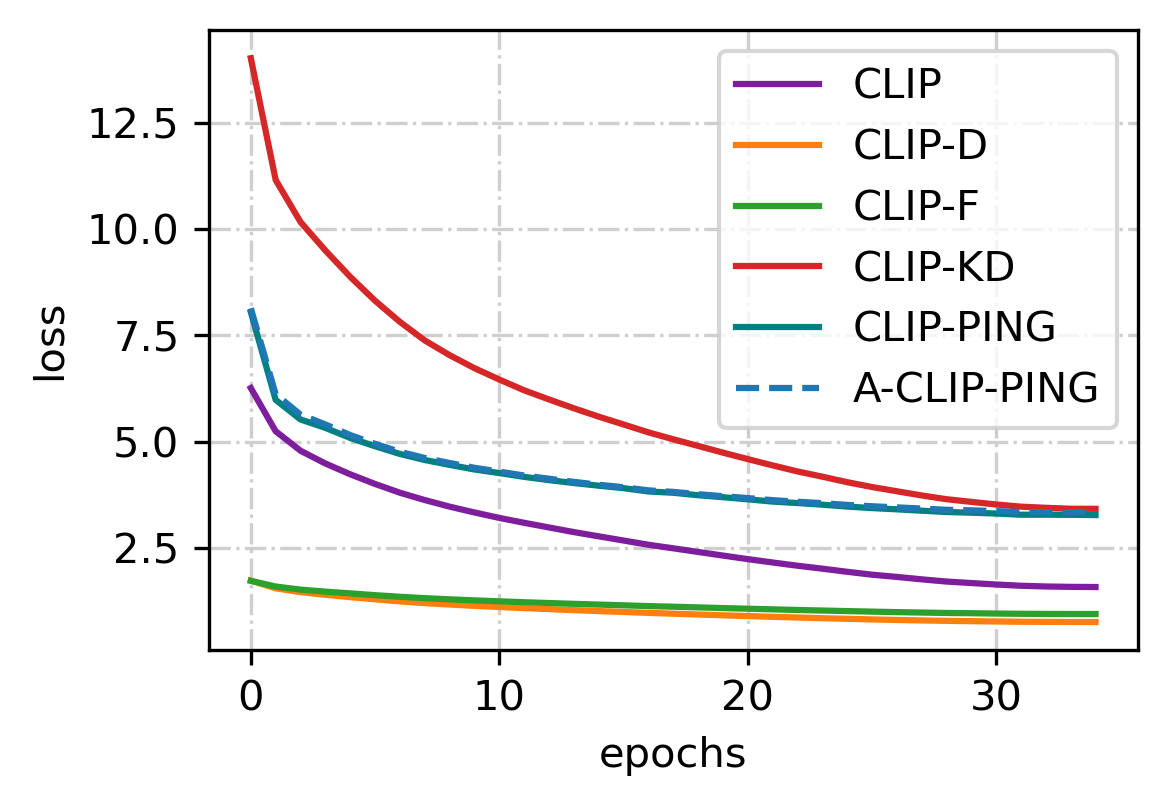}
  \caption{\footnotesize On COCO~\cite{lin2014microsoft}.}
  % \vspace{-0.5em}
  \end{subfigure}%
  ~
  \begin{subfigure}[t]{0.23\textwidth}
  \centering
  \includegraphics[width=\textwidth]{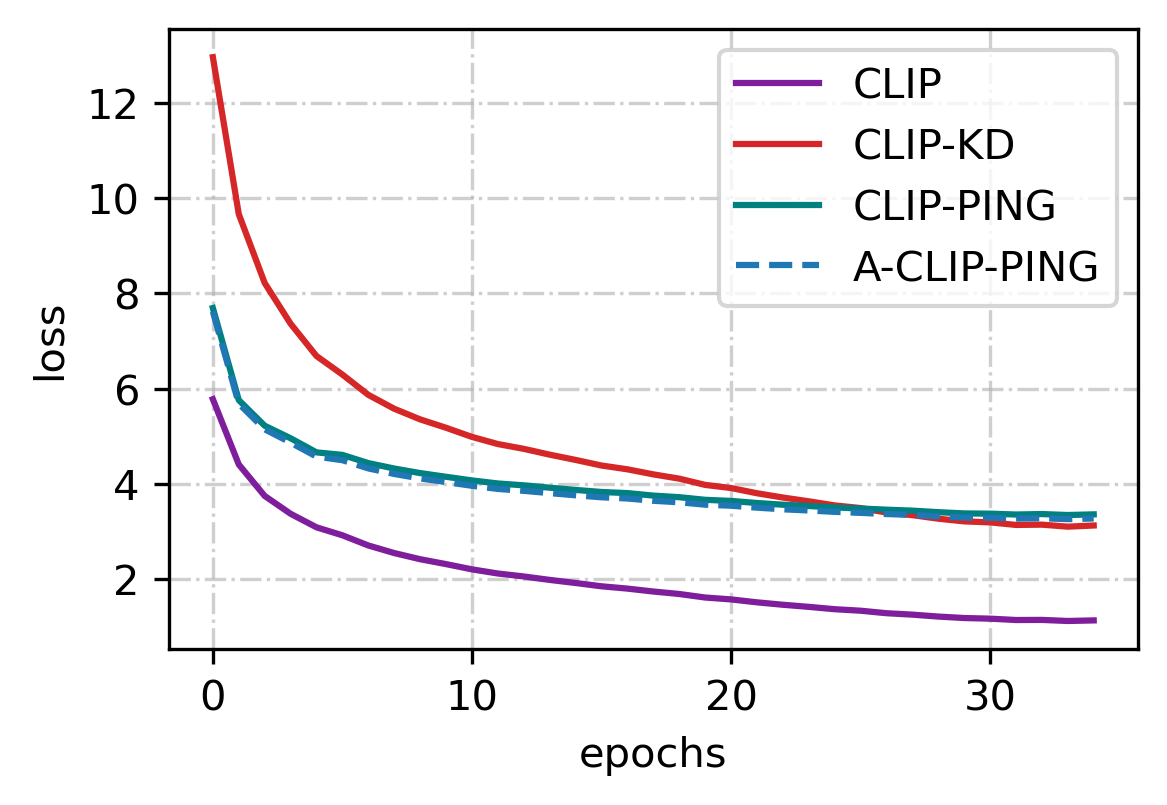}
  \caption{\footnotesize On COCO+CC3M~\cite{lin2014microsoft, sharma2018conceptual}}
  % \vspace{-0.5em}
  \end{subfigure}
   \caption{Curves for ViT-XS~\cite{dosovitskiy2020image} + MoblileBERT\textsubscript{TINY}~\cite{sun2020mobilebert}.}
   \label{fig:training_curves}
\end{figure}

\paragraph{Training convergence analysis}
We present the training curves of \mbox{CLIP-PING} and other baselines in Fig.~\ref{fig:training_curves} to illustrate the convergence behavior~\cite{beyer2022knowledge} of ViT-XS~\cite{dosovitskiy2020image} + MobileBERT\textsubscript{TINY}~\cite{sun2020mobilebert} across different methods.

\begin{table}[t]
\centering
\caption{Comparison on zero-shot classification performance. ViT-XS~\cite{dosovitskiy2020image} is the image encoder. The best results are marked in \textbf{bold}.}
\label{tab:zs}
\resizebox{\linewidth}{!}{
\begin{tabular}{lcccccc}
\toprule
\multicolumn{1}{c|}{Method}             & STL~\cite{coates2011analysis}    & C10~\cite{krizhevsky2009learning}      & C100~\cite{krizhevsky2009learning}      & SA-40~\cite{yao2011human} & \multicolumn{1}{c|}{IN-1K~\cite{deng2009imagenet}} & AVG  \\ \midrule
\multicolumn{7}{c}{Pre-training Dataset: COCO~\cite{lin2014microsoft} (600K)}                                                               \\ \midrule
\multicolumn{1}{l|}{CLIP}               & 66.7   & 25.1     & 7.7       & 32.2  & \multicolumn{1}{c|}{-}     & 32.9 \\
\multicolumn{1}{l|}{CLIP-D}             & \textbf{71.4}   & \textbf{41.8}     & 12.1      & 38.3  & \multicolumn{1}{c|}{-}     & 40.9 \\
\multicolumn{1}{l|}{CLIP-F}             & 69.2   & 40.2     & 11.7      & 36.1  & \multicolumn{1}{c|}{-}     & 39.3 \\
\multicolumn{1}{l|}{CLIP-KD}            & 64.9   & 26.3     & 7.5       & 32.2  & \multicolumn{1}{c|}{-}     & 32.7 \\
\multicolumn{1}{l|}{CLIP-PING (ours)}   & \textbf{71.4}   & 41.6     & \textbf{13.0}      & \textbf{38.6}  & \multicolumn{1}{c|}{-}     & \textbf{41.2} \\
\multicolumn{1}{l|}{A-CLIP-PING (ours)} & \textbf{75.1}   & \textbf{50.1}     & \textbf{17.3}      & \textbf{41.2}  & \multicolumn{1}{c|}{-}     & \textbf{45.9} \\ \midrule
\multicolumn{7}{c}{Pre-training Dataset: COCO+CC3M~\cite{lin2014microsoft, sharma2018conceptual} (3M)}                                                          \\ \midrule
\multicolumn{1}{l|}{CLIP}               & 82.7   & 59.5     & 24.6      & 52.3  & \multicolumn{1}{c|}{17.2}  & 47.3 \\
\multicolumn{1}{l|}{CLIP-KD}            & 83.9   & 64.8     & 25.7      & 57.3  & \multicolumn{1}{c|}{18.6}  & 50.1 \\
\multicolumn{1}{l|}{CLIP-PING (ours)}   & \textbf{87.1}   & \textbf{71.9}     & \textbf{31.3}      & \textbf{61.1}  & \multicolumn{1}{c|}{\textbf{22.7}}  & \textbf{54.8} \\
\multicolumn{1}{l|}{A-CLIP-PING (ours)} & \textbf{89.4}   & \textbf{76.9}     & \textbf{42.0}      & \textbf{63.7}  & \multicolumn{1}{c|}{\textbf{24.1}}  & \textbf{59.2} \\ \bottomrule
\end{tabular}
}
\end{table}

\begin{table}[t]
\centering
\caption{Comparison on zero-shot robustness evaluation performance. ViT-XS~\cite{dosovitskiy2020image} is the image encoder. The pre-training dataset is COCO+CC3M~\cite{lin2014microsoft, sharma2018conceptual} (3M). The best results are marked in \textbf{bold}.}
\label{tab:robustness}
\resizebox{\linewidth}{!}{
\begin{tabular}{l|cccc}
\toprule
\multicolumn{1}{c|}{Method} & IN-V2~\cite{recht2019imagenet}         & IN-R~\cite{hendrycks2021many}          & IN-O~\cite{hendrycks2021natural}          & IN-S~\cite{wang2019learning}          \\ \midrule
CLIP                        & 14.4          & 14.1          & 23.4          & 4.8           \\
CLIP-KD                     & 15.4          & 15.2          & 24.4          & 5.5           \\
CLIP-PING (ours)            & \textbf{19.5} & \textbf{19.7} & \textbf{29.0} & \textbf{8.7}  \\
A-CLIP-PING (ours)          & \textbf{20.3} & \textbf{20.7} & \textbf{31.0} & \textbf{10.1} \\ \bottomrule
\end{tabular}
}
\vspace{-1em}
\end{table}

\begin{table*}[t]
\centering
\caption{Comparison on linear evaluation performance. ViT-XS~\cite{dosovitskiy2020image} is the image encoder. Best results are marked in~\textbf{bold}.}
\label{tab:linear}
\resizebox{\linewidth}{!}{
\begin{tabular}{lccccccccccccr}
\toprule
\multicolumn{1}{c|}{\multirow{2}{*}{Method}} & \multicolumn{4}{c|}{Mean Per Class Accuracy}                                       & \multicolumn{8}{c|}{Accuracy}                                                                                                                      & \multicolumn{1}{c}{\multirow{2}{*}{AVG}} \\ \cmidrule{2-13}
\multicolumn{1}{c|}{}                        & Pets~\cite{parkhi2012cats}          & Cal.~\cite{fei2004learning}          & Flow.~\cite{nilsback2008automated}         & \multicolumn{1}{c|}{FGVC~\cite{maji2013fine}}          & Food~\cite{bossard2014food}          & DTD~\cite{cimpoi2014describing}           & SUN~\cite{xiao2010sun}           & Cars~\cite{krause2013collecting, krause20133d}          & STL~\cite{coates2011analysis}           & C10~\cite{krizhevsky2009learning}           & C100~\cite{krizhevsky2009learning}          & \multicolumn{1}{c|}{IN-1K~\cite{deng2009imagenet}}         & \multicolumn{1}{c}{}                     \\ \midrule
\multicolumn{14}{c}{Pre-training Dataset: COCO~\cite{lin2014microsoft} (600K)}                                                                                                                                                                                                                                                                            \\ \midrule
\multicolumn{1}{l|}{CLIP}                    & 42.5          & 58.3          & 61.5          & \multicolumn{1}{c|}{19.3}          & 47.6          & 43.3          & 49.9          & 14.2          & 80.6          & 66.3          & 39.8          & \multicolumn{1}{c|}{-}             & 47.6                                     \\
\multicolumn{1}{l|}{CLIP-D}                  & 47.5          & \textbf{68.0} & 66.7          & \multicolumn{1}{c|}{20.1}          & 52.1          & \textbf{48.2} & 55.6          & \textbf{17.9} & 83.2          & 68.0          & 45.8          & \multicolumn{1}{c|}{-}             & 52.1                                     \\
\multicolumn{1}{l|}{CLIP-F}                  & 46.5          & 62.6          & 65.0          & \multicolumn{1}{c|}{20.9}          & 48.9          & 45.2          & 52.2          & 17.5          & 82.4          & 67.6          & 43.0          & \multicolumn{1}{c|}{-}             & 50.2                                     \\
\multicolumn{1}{l|}{CLIP-KD}                 & 46.2          & 61.0          & 65.0          & \multicolumn{1}{c|}{20.8}          & 49.3          & 43.6          & 51.0          & 16.5          & 81.7          & 66.7          & 42.4          & \multicolumn{1}{c|}{-}             & 49.5                                     \\
\multicolumn{1}{l|}{CLIP-PING (ours)}        & \textbf{54.1} & 67.1          & \textbf{66.8} & \multicolumn{1}{c|}{\textbf{23.1}} & \textbf{54.7} & 47.3          & \textbf{56.1} & 17.8          & \textbf{84.9} & \textbf{69.6} & \textbf{45.9} & \multicolumn{1}{c|}{-}             & \textbf{53.4}                            \\
\multicolumn{1}{l|}{A-CLIP-PING (ours)}      & \textbf{58.6} & \textbf{70.7} & \textbf{69.4} & \multicolumn{1}{c|}{\textbf{25.7}} & \textbf{57.7} & \textbf{50.2} & \textbf{58.4} & \textbf{21.0} & \textbf{87.9} & \textbf{75.3} & \textbf{50.3} & \multicolumn{1}{c|}{-}             & \textbf{56.8}                            \\ \midrule
\multicolumn{14}{c}{Pre-training Dataset: COCO+CC3M~\cite{lin2014microsoft, sharma2018conceptual} (3M)}                                                                                                                                                                                                                                                                         \\ \midrule
\multicolumn{1}{l|}{CLIP}                    & 52.4          & 73.7          & 71.1          & \multicolumn{1}{c|}{23.7}          & 57.0          & 58.7          & 65.1          & 19.3          & 90.2          & 75.2          & 50.5          & \multicolumn{1}{c|}{46.5}          & 56.2                                     \\
\multicolumn{1}{l|}{CLIP-KD}                 & 61.5          & 79.3          & 79.1          & \multicolumn{1}{c|}{26.3}          & 59.6          & 60.3          & 66.2          & 23.8          & 91.7          & 75.2          & 52.9          & \multicolumn{1}{c|}{50.2}          & 60.5                                     \\
\multicolumn{1}{l|}{CLIP-PING (ours)}        & \textbf{74.2} & \textbf{84.2} & \textbf{81.9} & \multicolumn{1}{c|}{\textbf{29.4}} & \textbf{63.9} & \textbf{63.2} & \textbf{68.7} & \textbf{29.6} & \textbf{94.1} & \textbf{80.4} & \textbf{58.1} & \multicolumn{1}{c|}{\textbf{55.7}} & \textbf{65.3}                            \\
\multicolumn{1}{l|}{A-CLIP-PING (ours)}      & \textbf{77.5} & \textbf{86.3} & \textbf{83.4} & \multicolumn{1}{c|}{\textbf{30.8}} & \textbf{65.3} & \textbf{65.1} & \textbf{69.6} & \textbf{30.8} & \textbf{95.6} & \textbf{86.7} & \textbf{67.2} & \multicolumn{1}{c|}{\textbf{58.4}} & \textbf{68.1}                            \\ \bottomrule
\end{tabular}
}
\end{table*}

\begin{table*}[t]
\centering
\caption{Effect of model scaling: {\color{gray}CLIP} vs. CLIP-PING. Memory usage is recorded on a single NVIDIA A100 GPU.}
\label{tab:model_scale}
\resizebox{\linewidth}{!}{
\begin{tabular}{l|c|c|c|cc|cc|cc|cc}
\toprule
\multicolumn{1}{c|}{\multirow{2}{*}{\begin{tabular}[c]{@{}c@{}}Image\\ Encoder\end{tabular}}} & \multirow{2}{*}{\begin{tabular}[c]{@{}c@{}}Text\\ Encoder\end{tabular}}          & \multirow{2}{*}{\begin{tabular}[c]{@{}c@{}}Total\\ \#Params\end{tabular}} & \multirow{2}{*}{\begin{tabular}[c]{@{}c@{}}Memory\\ (MiB)\end{tabular}} & \multicolumn{2}{c|}{CC3M~\cite{sharma2018conceptual}} & \multicolumn{2}{c|}{COCO~\cite{lin2014microsoft}} & \multicolumn{2}{c|}{Flickr30K~\cite{young2014image}} & \multicolumn{2}{c}{IN-1K~\cite{deng2009imagenet}} \\
\multicolumn{1}{c|}{}                                                               &                                                                                  &                                                                           &                                                                         & I2T@1         & T2I@1         & I2T@1         & T2I@1         & I2T@1            & T2I@1           & LE          & ZS          \\ \bottomrule
{\color{gray}ViT-M~\cite{dosovitskiy2020image} (CLIP)}                                                                 & \multirow{4}{*}{\begin{tabular}[c]{@{}c@{}}MobileBERT~\cite{sun2020mobilebert}\\ (25.2M)\end{tabular}} & {\color{gray}63.9M}                                                       & {\color{gray}13684}                                 & {\color{gray}27.3}       & {\color{gray}27.9}        & {\color{gray}37.7}        & {\color{gray}26.1}        & {\color{gray}47.3}           & {\color{gray}34.0}          & {\color{gray}52.3}        & {\color{gray}19.4}        \\ \cmidrule{1-1} \cmidrule{3-12} 
ViT-M~\cite{dosovitskiy2020image}                                                                               &                                                                                  & 63.9M                                                                     & 15544                                               & 32.1                     & 32.0                      & 41.6                      & 29.9        & 51.6           & 40.4          & 64.2        & 27.5        \\
ConvNeXt-Tiny~\cite{liu2022convnet}                                                                       &                                                                                  & 53.8M                                                                     & 26170                                               & 29.9                     & 31.0                      & 43.4                      & 30.9        & 57.8           & 41.6          & 65.3        & 28.2        \\
MNv4-Hybrid-L~\cite{qin2024mobilenetv4}                                                                       &                                                                                  & 63.6M                                                                     & 19196                                               & 30.3                     & 30.9                      & 41.2                      & 30.0        & 55.5           & 40.7          & 63.3        & 27.9        \\ \bottomrule
\end{tabular}
}
\end{table*}

\paragraph{Model scaling analysis}
Our primary focus is on resource-constrained scenarios, which motivates us to prioritize computational efficiency over large-scale data and models.
Yet, to further validate the performance consistency of \mbox{CLIP-PING} within our constraints, we provide experiments with diverse mid-scale models trained on the COCO+CC3M~\cite{lin2014microsoft, sharma2018conceptual} dataset in Table~\ref{tab:model_scale}.
We consider MobileBERT~\cite{sun2020mobilebert} (with MobileBertTokenizer and a maximum context length of 55) as our mid-scale text encoder. For the image encoders, we employ: ViT-M~\cite{dosovitskiy2020image} (a mid-scale transformer-based model), ConvNeXt-Tiny~\cite{liu2022convnet} (a mid-scale convolutional network), and MNv4-Hybrid-L~\cite{qin2024mobilenetv4} (a mid-scale hybrid convolutional transformer variant of MobileNetv4 model).
This analysis highlights CLIP-PING's consistent improvements with model scales across diverse variations of image encoder architectures.

\begin{table}[t]
\centering
\caption{Influence of loss weights $\alpha$ and $\lambda$ on retrieval performance for Flickr30K~\cite{young2014image} and average zero-shot top-1 classification accuracy across 4 datasets, using ViT-XS~\cite{dosovitskiy2020image} image encoder pre-trained on COCO~\cite{lin2014microsoft} dataset.}
\label{tab:alpha_lambda}
\begin{subtable}[t]{0.22\textwidth}
\centering
\caption{\footnotesize The $\alpha$ effect.}
\label{tab:alpha}
\resizebox{\linewidth}{!}{
\begin{tabular}{c|cc|c}
\toprule
\multirow{2}{*}{$\alpha$} & \multicolumn{2}{c|}{Flickr30K~\cite{young2014image}} & ZS   \\
                       & I2T@1          & T2I@1         & AVG  \\ \midrule
0                      & 26.4           & 18.9          & 38.0 \\
0.25                   & \textbf{28.1}           & \textbf{20.2}          & \textbf{41.2} \\
0.5                    & 27.4           & 19.2          & 40.0 \\
0.75                   & 24.9           & 18.4          & 36.7 \\
1                      & 20.6           & 16.5          & 35.1 \\ \bottomrule
\end{tabular}
}
\end{subtable}
\hspace{\fill}
\begin{subtable}[t]{0.24\textwidth}
\centering
\caption{\footnotesize The $\lambda$ effect.}
\label{tab:lambda}
\resizebox{\linewidth}{!}{
\begin{tabular}{c|cc|c}
\toprule
\multirow{2}{*}{$\lambda$} & \multicolumn{2}{c|}{Flickr30K~\cite{young2014image}} & ZS   \\
                       & I2T@1          & T2I@1         & AVG  \\ \midrule
0.2                    & 23.3           & 16.9          & 35.5 \\
0.4                    & 24.9           & 18.7          & 39.0 \\
0.6                    & \textbf{28.1}           & \textbf{20.2}          & \textbf{41.2} \\
0.8                    & 24.5           & 17.2          & 40.8 \\ \bottomrule
\end{tabular}
}
\end{subtable}
\end{table}

\begin{table}
\centering
\caption{Impact of support set size $|Q|$ and top-$k$ neighbors on retrieval performance for Flickr30K~\cite{young2014image} and average zero-shot top-1 classification accuracy across 4 datasets, using ViT-XS~\cite{dosovitskiy2020image} image encoder pre-trained on COCO~\cite{lin2014microsoft} dataset.}
\label{tab:q_k_d_i}
\begin{subtable}[t]{0.24\textwidth}
\centering
\caption{\footnotesize Support set size.}
\label{tab:queue_size}
\resizebox{\linewidth}{!}{
\begin{tabular}{c|cc|c}
\toprule
\multirow{2}{*}{$|Q|$} & \multicolumn{2}{c|}{Flickr30K~\cite{young2014image}} & ZS   \\
                       & I2T@1          & T2I@1         & AVG  \\ \midrule
8192                   & 25.4           & 18.9          & 39.3 \\
16384                  & 27.0           & 19.7          & 40.2 \\
32768                  & \textbf{28.1}           & \textbf{20.2}          & \textbf{41.2} \\
65536                  & 25.2           & 20.0          & 40.9 \\
98304                  & 24.3           & 19.1          & 40.4 \\ \bottomrule
\end{tabular}
}
\end{subtable}
\hspace{\fill}
\begin{subtable}[t]{0.22\textwidth}
\centering
\caption{\footnotesize Top-$k$ neighbors.}
\label{tab:kNN}
\resizebox{\linewidth}{!}{
\begin{tabular}{c|cc|c}
\toprule
\multirow{2}{*}{$k$} & \multicolumn{2}{c|}{Flickr30K~\cite{young2014image}} & ZS   \\
                       & I2T@1          & T2I@1         & AVG  \\ \midrule
1                    & \textbf{28.1}           & \textbf{20.2}          & \textbf{41.2} \\
2                    & 25.5           & 19.2          & 40.6 \\
4                    & 26.2           & 19.3          & 38.2 \\
8                    & 24.6           & 18.7          & 39.5 \\
16                    & 26.7           & 18.1          & 38.7 \\ \bottomrule
\end{tabular}
}
\end{subtable}
\vspace{-1em}
\end{table}
 
\paragraph{Influence of $\alpha$}
As shown in Table~\ref{tab:alpha}, we explore the effect of supervision loss weight $\alpha$ in Eq.~(\ref{eq:ping}).
As $\alpha$ increases, emphasizing more weight on XNN supervision, the performance gradually declines.
However, the drop in performance saturates at $\alpha=0.25$, suggesting that \mbox{CLIP-PING} benefits both from NN and XNN supervision.
This indicates that both forms of supervision contribute positively, with $\alpha = 0.25$ being the optimal choice to balance them.

\paragraph{Influence of $\lambda$}
As shown in Table~\ref{tab:lambda}, we examine the effect of loss weight $\lambda$ in Eq.~(\ref{eq:final_loss}).
Performance gradually improves with increasing $\lambda$ up to 0.6, after which it plateaus, indicating that intrinsic neighbors guidance boosts the training while \mbox{CLIP-PING} continues to benefit from foundational cross-modal contrastive alignment. 

\paragraph{Impact of support set size}
Table~\ref{tab:queue_size} presents the results of varying the support set size $|Q|$.
As the support set grows, it increases the chance of retrieving a closer NN within the dataset.
However, increasing the support set size beyond 32768 does not yield additional performance gains.

\begin{table}
\centering
\caption{Ablations of feature extractor and supervision source on retrieval performance for Flickr30K~\cite{young2014image} and average zero-shot top-1 classification accuracy across 4 datasets, using ViT-XS~\cite{dosovitskiy2020image} image encoder pre-trained on COCO~\cite{lin2014microsoft} dataset. The best results are marked in \textbf{bold}.}
\label{tab:fex_nn}
\begin{subtable}[t]{0.215\textwidth}
\centering
\caption{\footnotesize The ViT-B/16~\cite{dosovitskiy2020image} effect.}
\label{tab:fex}
\resizebox{\linewidth}{!}{
\begin{tabular}{l|cc|c}
\toprule
\multicolumn{1}{c|}{\multirow{2}{*}{Method}} & \multicolumn{2}{c|}{Flickr30K~\cite{young2014image}} & ZS   \\
\multicolumn{1}{c|}{}                        & I2T@1          & T2I@1         & AVG  \\ \midrule
CLIP-D                                       & \textbf{25.5}           & 16.8          & 39.1  \\
CLIP-F                                       & 20.0           & 15.8          & 37.3  \\
CLIP-KD                                      & 21.8           & 15.3          & 33.5 \\
CLIP-PING                                    & 25.1           & \textbf{17.9}          & \textbf{39.2} \\
A-CLIP-PING                                  & \textbf{27.8}           & \textbf{19.4}          & \textbf{40.7} \\ \bottomrule
\end{tabular}
}
\end{subtable}
\hspace{\fill}
\begin{subtable}[t]{0.25\textwidth}
\centering
\caption{\footnotesize Supervision source.}
\label{tab:nn}
\resizebox{\linewidth}{!}{
\begin{tabular}{c|cc|c}
\toprule
\multirow{2}{*}{\begin{tabular}[c]{@{}c@{}}Super-\\ vision\end{tabular}} & \multicolumn{2}{c|}{Flickr30K~\cite{young2014image}} & ZS   \\
                             & I2T@1          & T2I@1         & AVG  \\ \midrule
both                         & \textbf{28.1}           & \textbf{20.2}          & \textbf{41.2} \\
txt only                      & 22.0           & 16.8          & 34.6 \\
img only                      & 23.6           & 17.7          & 38.8 \\ \bottomrule
\end{tabular}
}
\end{subtable}
\vspace{-1em}
\end{table}

\paragraph{Impact of Top-$k$ neighbors}
In Table~\ref{tab:kNN}, we explore the impact of selecting one random neighbor from the top-$k$ NN instead of the closest neighbor.
We find that increasing $k$ beyond 1 slightly degrades the performance, suggesting that the nearest-neighbor strategy offers clearer guidance, while other neighbors may introduce noise.

\paragraph{Impact of unimodal feature extractor}
In Table~\ref{tab:fex}, we employ ViT-B/16~\cite{dosovitskiy2020image}, pre-trained on ImageNet~\cite{deng2009imagenet} (i.e., vit\_base\_patch16\_224) as our image teacher or feature extractor and analyze its effect on performance.
Notably, our \mbox{CLIP-PING} variants still outperform the other methods.

\paragraph{Impact of supervision modality}
In Table~\ref{tab:nn}, we analyze the effect of supervision sources: both image and text supervision (default), with only text supervision (w/o $\mathcal{L}_I^{\text{NN}}$ and $\mathcal{L}_I^{\text{XNN}}$ in Eqs.~(\ref{eq:ping_nn}) and (\ref{eq:ping_xnn})), and with only image supervision (w/o $\mathcal{L}_T^{\text{NN}}$ and $\mathcal{L}_T^{\text{XNN}}$ in Eqs.~(\ref{eq:ping_nn}) and (\ref{eq:ping_xnn})).
We observed that \mbox{CLIP-PING} benefits equally from both supervision.

%-------------------------------------------------------------------------

\section{Discussion on limitations and future work}\label{sec:discussion}

The absence of explicit distillation in \mbox{CLIP-PING} may limit its ability to directly leverage real-time knowledge from larger pre-trained models.
To fill this gap, we introduced \mbox{A-CLIP-PING}, a variant that integrates teacher-student learning into the framework.
\mbox{A-CLIP-PING} achieves a compelling balance between computational efficiency and performance gains, making the choice between \mbox{CLIP-PING} and \mbox{A-CLIP-PING} dependent on the specific trade-offs required by different applications.
Our findings show that \mbox{CLIP-PING} excels in resource-constrained settings, offering an effective solution for efficient multi-modal learning.
However, its scalability for large-scale pre-training on excessively large multi-modal datasets remains unexplored due to computational limitations.
Our study aims to provide valuable insights and encourage further exploration into lightweight and efficient vision-language models.
Future research could extend \mbox{CLIP-PING} to additional modalities beyond image and text, as well as fine-tune it for diverse on-device tasks to further unlock its potential.

\section{Conclusion}\label{sec:conclusion}

In this work, we proposed \mbox{CLIP-PING}: Contrastive Language-Image Pre-training with Proximus Intrinsic Neighbors Guidance, an efficient training paradigm, to boost the performance of lightweight vision-language models.
By leveraging off-the-shelf pre-trained encoders and incorporating intra-modal and inter-modal supervision through frozen nearest neighbor samples, \mbox{CLIP-PING} significantly improves the representation learning capabilities of lightweight models, while reducing the computational and data demands.
Extensive experiments demonstrate the effectiveness of \mbox{CLIP-PING} in zero-shot classification and cross-modal retrieval tasks, achieving competitive performance even with limited resources.
Additionally, \mbox{CLIP-PING} exhibits strong transferability under linear evaluation protocol across several downstream tasks, making it a promising solution for efficient training of lightweight vision-language models in resource-constrained scenarios.

%-------------------------------------------------------------------------

\section{Additional implementation  details}\label{app:details}

\paragraph{Implementation of baselines}
For CLIP~\cite{radford2021learning}, we employ the standard image-text contrastive loss, i.e., Eq. (3).
\mbox{CLIP-D} and \mbox{CLIP-F} integrate a distillation loss~\cite{vasu2024mobileclip} into the standard contrastive loss with loss weight set to $\lambda=0.75$, as denoted by:
\begin{align}
    \mathcal{L}^{\text{CLIP-Distill}}&=(1-\lambda) \cdot \mathcal{L}^{\text{CLIP}}+\lambda \cdot \mathcal{L}^{\text{Distill}}, \\
    \mathcal{L}^{\text{Distill}}&=\frac{1}{2}(\mathcal{L}_{\text{Distill}}^{\text{I2T}}+\mathcal{L}_{\text{Distill}}^{\text{T2I}}),
      \label{eq:clip_distill}
\end{align}
where $\mathcal{L}_{\text{Distill}}^{\text{I2T}}$ and $\mathcal{L}_{\text{Distill}}^{\text{T2I}}$ are KL-divergence losses.
In \mbox{CLIP-F}, pre-computed frozen features, similar to those used in \mbox{CLIP-PING}, are utilized.
For CLIP-KD~\cite{yang2024clipkd}, we implement the objective function as specified in its original paper:
\begin{align}
    \mathcal{L}^{\text{CLIP-KD}}&=\mathcal{L}^{\text{CLIP}}+\lambda \cdot \mathcal{L}^{\text{KD}},
      \label{eq:clipkd}
\end{align}
where $\mathcal{L}^{\text{KD}}$ is the combination $\mathcal{L}^{\text{FD}}+\mathcal{L}^{\text{CRD}}+\mathcal{L}^{\text{ICL}}$ for Feature Distillation (FD), Contrastive Relational Distillation (CRD) and Interactive Contrastive Learning (ICL), with loss weights set to $\lambda_{FD}=2000$, and $\lambda_{CRD}=\lambda_{ICL}=1$.

\begin{table*}[b]
\centering
\caption{Comparison on memory usage and training duration per epoch. Experiments for COCO~\cite{lin2014microsoft} dataset are run on a single NVIDIA RTX A6000 GPU, while those for COCO+CC3M~\cite{lin2014microsoft, sharma2018conceptual} dataset are run on a single NVIDIA A100 GPU.}
\label{tab:mem_time}
\begin{subtable}[t]{0.32\textwidth}
\centering
\caption{\footnotesize With ViT-XS~\cite{dosovitskiy2020image} image encoder.}
\label{tab:mem_time_vit}
\resizebox{\linewidth}{!}{
\begin{tabular}{l|cc|cc}
\toprule
\multicolumn{1}{c|}{\multirow{2}{*}{Method}} & \multicolumn{2}{c|}{COCO (600K)}                                                                                                            & \multicolumn{2}{c}{COCO+CC3M (3M)}                                                                                                        \\ \cmidrule{2-5} 
\multicolumn{1}{c|}{}                        & \multicolumn{1}{c|}{\begin{tabular}[c]{@{}c@{}}Memory\\ (MiB)$\downarrow$\end{tabular}} & \begin{tabular}[c]{@{}c@{}}Time\\ (hours)$\downarrow$\end{tabular} & \multicolumn{1}{c|}{\begin{tabular}[c]{@{}c@{}}Memory\\ (MiB)$\downarrow$\end{tabular}} & \begin{tabular}[c]{@{}c@{}}Time\\ (hours)$\downarrow$\end{tabular} \\ \midrule
CLIP                                         & \multicolumn{1}{c|}{11074}                                                  & 0.19                                                   & \multicolumn{1}{c|}{11379}                                                  & 0.86                                                   \\
CLIP-KD                                      & \multicolumn{1}{c|}{25036}                                                  & 0.35                                                   & \multicolumn{1}{c|}{25305}                                                  & 1.18                                                   \\
CLIP-PING                                    & \multicolumn{1}{c|}{11580}                                                  & 0.19                                                   & \multicolumn{1}{c|}{11885}                                                  & 0.88                                                   \\
A-CLIP-PING                                  & \multicolumn{1}{c|}{25370}                                                  & 0.35                                                   & \multicolumn{1}{c|}{25639}                                                  & 1.15                                                   \\ \bottomrule
\end{tabular}
}
\end{subtable}
\hspace{\fill}
\begin{subtable}[t]{0.32\textwidth}
\centering
\caption{\footnotesize With ConvNeXt-Pico~\cite{liu2022convnet} image encoder.}
\label{tab:mem_time_convnext}
\resizebox{\linewidth}{!}{
\begin{tabular}{l|cc|cc}
\toprule
\multicolumn{1}{c|}{\multirow{2}{*}{Method}} & \multicolumn{2}{c|}{COCO (600K)}                                                                                                            & \multicolumn{2}{c}{COCO+CC3M (3M)}                                                                                                        \\ \cmidrule{2-5} 
\multicolumn{1}{c|}{}                        & \multicolumn{1}{c|}{\begin{tabular}[c]{@{}c@{}}Memory\\ (MiB)$\downarrow$\end{tabular}} & \begin{tabular}[c]{@{}c@{}}Time\\ (hours)$\downarrow$\end{tabular} & \multicolumn{1}{c|}{\begin{tabular}[c]{@{}c@{}}Memory\\ (MiB)$\downarrow$\end{tabular}} & \begin{tabular}[c]{@{}c@{}}Time\\ (hours)$\downarrow$\end{tabular} \\ \midrule
CLIP                                         & \multicolumn{1}{c|}{16126}                                                  & 0.24                                                   & \multicolumn{1}{c|}{16299}                                                  & 1.08                                                   \\
CLIP-KD                                      & \multicolumn{1}{c|}{28460}                                                  & 0.41                                                   & \multicolumn{1}{c|}{28633}                                                  & 1.42                                                   \\
CLIP-PING                                    & \multicolumn{1}{c|}{16874}                                                  & 0.25                                                   & \multicolumn{1}{c|}{18615}                                                  & 1.13                                                   \\
A-CLIP-PING                                  & \multicolumn{1}{c|}{28792}                                                  & 0.41                                                   & \multicolumn{1}{c|}{28965}                                                  & 1.39                                                   \\ \bottomrule
\end{tabular}
}
\end{subtable}
\hspace{\fill}
\begin{subtable}[t]{0.32\textwidth}
\centering
\caption{\footnotesize With MNv4-Hybrid-M~\cite{qin2024mobilenetv4} image encoder.}
\label{tab:mem_time_mnv4}
\resizebox{\linewidth}{!}{
\begin{tabular}{l|cc|cc}
\toprule
\multicolumn{1}{c|}{\multirow{2}{*}{Method}} & \multicolumn{2}{c|}{COCO (600K)}                                                                                                            & \multicolumn{2}{c}{COCO+CC3M (3M)}                                                                                                        \\ \cmidrule{2-5} 
\multicolumn{1}{c|}{}                        & \multicolumn{1}{c|}{\begin{tabular}[c]{@{}c@{}}Memory\\ (MiB)$\downarrow$\end{tabular}} & \begin{tabular}[c]{@{}c@{}}Time\\ (hours)$\downarrow$\end{tabular} & \multicolumn{1}{c|}{\begin{tabular}[c]{@{}c@{}}Memory\\ (MiB)$\downarrow$\end{tabular}} & \begin{tabular}[c]{@{}c@{}}Time\\ (hours)$\downarrow$\end{tabular} \\ \midrule
CLIP                                         & \multicolumn{1}{c|}{16696}                                                  & 0.28                                                   & \multicolumn{1}{c|}{15747}                                                  & 1.21                                                   \\
CLIP-KD                                      & \multicolumn{1}{c|}{31780}                                                  & 0.44                                                   & \multicolumn{1}{c|}{30831}                                                  & 1.54                                                   \\
CLIP-PING                                    & \multicolumn{1}{c|}{17050}                                                  & 0.28                                                   & \multicolumn{1}{c|}{16101}                                                  & 1.24                                                   \\
A-CLIP-PING                                  & \multicolumn{1}{c|}{32110}                                                  & 0.44                                                   & \multicolumn{1}{c|}{31161}                                                  & 1.52                                                   \\ \bottomrule
\end{tabular}
}
\end{subtable}
\end{table*}

\paragraph{Additional baseline}
Following the official repository, we implemented DeCLIP~\cite{li2022supervision} and trained it on the COCO~\cite{lin2014microsoft} dataset.
Since DeCLIP~\cite{li2022supervision} requires encoding each (image, text) pair twice and relies heavily on data augmentation, it incurs significantly higher computational costs.
To manage GPU memory constraints, we reduced the batch size by half, from 1024 to 512.
Additionally, due to training instability, we adjusted the learning rates to 1.5e-3 for the image encoder and 2.5e-4 for the text encoder.
Training DeCLIP~\cite{li2022supervision} with the MNv4-Hybrid-M~\cite{qin2024mobilenetv4} image encoder on the COCO~\cite{lin2014microsoft} (600K) dataset for 35 epochs required approximately 24 hours on a single NVIDIA RTX A6000 GPU --- \textbf{2.4$\times$} longer than \mbox{CLIP-PING} and \textbf{1.5$\times$} longer than \mbox{A-CLIP-PING}.
Given the prolonged training duration and suboptimal performance on cross-modal retrieval tasks, we limited our experiments with DeCLIP~\cite{li2022supervision}.
Notably, while DeCLIP~\cite{li2022supervision} is designed to learn visual representations through the use of broader and scalable supervision in data-efficient settings, its performance with lightweight models in our evaluations fell short of \mbox{CLIP-PING}.
Table~\ref{tab:declip} summarizes the results, highlighting the superior performance and efficiency of \mbox{CLIP-PING} over DeCLIP~\cite{li2022supervision}.

\begin{table}[h]
\centering
\caption{Comparison with DeCLIP~\cite{li2022supervision} on cross-modal retrieval tasks and average zero-shot top-1 accuracy across 4 datasets. DeCLIP is trained with a batch size of 512. The pre-training dataset is COCO~\cite{lin2014microsoft} (600K). The best results are marked in \textbf{bold}. Memory usage and wall-clock time per epoch are measured on a single NVIDIA RTX A6000 GPU.}
\label{tab:declip}
\resizebox{\linewidth}{!}{
\begin{tabular}{lccccccc}
\toprule
\multicolumn{1}{c|}{\multirow{2}{*}{Method}} & \multicolumn{1}{c|}{\multirow{2}{*}{\begin{tabular}[c]{@{}c@{}}Memory\\ (MiB)$\downarrow$\end{tabular}}} & \multicolumn{1}{c|}{\multirow{2}{*}{\begin{tabular}[c]{@{}c@{}}Time\\ (hours)$\downarrow$\end{tabular}}} & \multicolumn{2}{c|}{COCO~\cite{lin2014microsoft}}          & \multicolumn{2}{c|}{Flickr30K~\cite{young2014image}}     & \multirow{2}{*}{\begin{tabular}[c]{@{}c@{}}ZS\\ AVG\end{tabular}} \\
\multicolumn{1}{c|}{}                        & \multicolumn{1}{c|}{}               & \multicolumn{1}{c|}{}                                                         & I2T@1 & \multicolumn{1}{c|}{T2I@1} & I2T@1 & \multicolumn{1}{c|}{T2I@1} &                                                                   \\ \midrule
\multicolumn{8}{c}{Model: ViT-XS~\cite{dosovitskiy2020image} + MoblileBERT\textsubscript{TINY}~\cite{sun2020mobilebert}}                                                                                                                                                                                                                                      \\ \midrule
\multicolumn{1}{l|}{DeCLIP\textsubscript{512}}                  & \multicolumn{1}{c|}{27080}                 & \multicolumn{1}{c|}{0.52}                                             & 7.8   & \multicolumn{1}{c|}{6.5}   & 9.0   & \multicolumn{1}{c|}{7.1}   & 36.7                                                              \\
\multicolumn{1}{l|}{CLIP-PING (ours)}        & \multicolumn{1}{c|}{11580}                                    & \multicolumn{1}{c|}{0.19}                               & \textbf{24.7}  & \multicolumn{1}{c|}{\textbf{18.4}}  & \textbf{28.1}  & \multicolumn{1}{c|}{\textbf{20.2}}  & \textbf{41.2}                                                              \\ \midrule
\multicolumn{8}{c}{Model: ConvNeXt-Pico~\cite{liu2022convnet} + MoblileBERT\textsubscript{TINY}~\cite{sun2020mobilebert}}                                                                                                                                                                                                                               \\ \midrule
\multicolumn{1}{l|}{DeCLIP\textsubscript{512}}                  & \multicolumn{1}{c|}{31362}                  & \multicolumn{1}{c|}{0.63}                                            & 8.7   & \multicolumn{1}{c|}{7.8}   & 7.2   & \multicolumn{1}{c|}{7.1}   & 38.2                                                              \\
\multicolumn{1}{l|}{CLIP-PING (ours)}        & \multicolumn{1}{c|}{16874}                                     & \multicolumn{1}{c|}{0.25}                              & \textbf{27.3}  & \multicolumn{1}{c|}{\textbf{20.3}}  & \textbf{29.6}  & \multicolumn{1}{c|}{\textbf{20.8}}  & \textbf{41.8}                                                              \\ \midrule
\multicolumn{8}{c}{Model: MNv4-Hybrid-M~\cite{qin2024mobilenetv4} + MoblileBERT\textsubscript{TINY}~\cite{sun2020mobilebert}}                                                                                                                                                                                                                               \\ \midrule
\multicolumn{1}{l|}{DeCLIP\textsubscript{512}}                  & \multicolumn{1}{c|}{31320}                   & \multicolumn{1}{c|}{0.68}                                           & 11.2  & \multicolumn{1}{c|}{9.8}   & 10.3  & \multicolumn{1}{c|}{8.7}   & 39.0                                                              \\
\multicolumn{1}{l|}{CLIP-PING (ours)}        & \multicolumn{1}{c|}{17050}                                      & \multicolumn{1}{c|}{0.28}                             & \textbf{27.8}  & \multicolumn{1}{c|}{\textbf{21.1}}  & \textbf{29.9}  & \multicolumn{1}{c|}{\textbf{21.5}}  & \textbf{42.3}                                                              \\ \bottomrule
\end{tabular}
}
\vspace{-1em}
\end{table}

\paragraph{Training efficiency}
Noting that no single indicator is sufficient for measuring efficiency~\cite{dehghani2022the}, Table~\ref{tab:mem_time} reports both memory usage and wall-clock times for training one epoch with a batch size of 1024.
Each epoch comprises 577 iterations for the COCO~\cite{lin2014microsoft} (600K) dataset and 2,799 iterations for the COCO+CC3M~\cite{lin2014microsoft, sharma2018conceptual} (3M) dataset.
While CLIP-KD~\cite{yang2024clipkd} is approximately \textbf{1.3$\times$} to \textbf{1.8$\times$} slower than standard CLIP training, \mbox{CLIP-PING} maintains a comparable efficiency.
Additionally, \mbox{A-CLIP-PING} strikes a compelling balance between computational cost and performance gains.

\begin{algorithm*}[t]
    \caption{CLIP-PING}
    \begin{algorithmic}[1]
        \renewcommand{\algorithmicrequire}{\textbf{Input:}}
        \renewcommand{\algorithmicensure}{\textbf{Output:}}
        \REQUIRE {pre-trained encoders $\mathcal{F}_I^*$ and $\mathcal{F}_T^*$, dataset $\mathcal{D}=\{(I_k,T_k)\}_{k=1}^{|\mathcal{D}|}$}, loss weights $(\alpha, \lambda)$, batch size $N$
        \ENSURE lightweight vision-language model $(\mathcal{E}_I$, $\mathcal{E}_T)$ \\
        
        \vspace{0.5em}
        \textcolor{gray}{*****/*** Unimodal Feature Extraction ***/*****} \\
        \STATE \textbf{Initialize:} auxiliary feature banks $\mathcal{B}_I^*$ and $\mathcal{B}_T^*$
        \FOR{$k$ in range($|D|$)}
            \STATE $\mathcal{B}_I^* \gets \tilde{z}_k^I$, where $\tilde{z}_k^I=\mathcal{F}_I^*(I_k)$ \hspace*{\fill} \textcolor{gray}{\# extract image features and store them frozen in image auxiliary feature bank}
            \vspace{0.2em}
            \STATE $\mathcal{B}_T^* \gets \tilde{z}_k^T$, where $\tilde{z}_k^T=\mathcal{F}_T^*(T_k)$ \hspace*{\fill} \textcolor{gray}{\# extract text features and store them frozen in text auxiliary feature bank}
        \ENDFOR \\
        
        \vspace{0.5em}
        \textcolor{gray}{*****/*** Multi-modal Training ***/*****} \\
        \STATE \textbf{Initialize:} lightweight vision-language model $(\mathcal{E}_I$, $\mathcal{E}_T)$, support sets $\mathcal{Q}_I \subset \mathcal{B}_I^*$ and $\mathcal{Q}_T \subset \mathcal{B}_T^*$
        \WHILE{training}
        % \FOR{$e$ in range($E$)}
            \FOR{$k$ in range($N$)}
                \STATE $z_k^I=\mathcal{E}_I(I_k)$, \quad $\tilde{z}_k^I \gets \mathcal{B}_I^*(I_k)$ \hspace*{\fill} \textcolor{gray}{\# extract image features and get corresponding frozen features}
                \vspace{0.2em}
                \STATE $z_k^T=\mathcal{E}_T(T_k)$, \quad $\tilde{z}_k^T \gets \mathcal{B}_T^*(T_k)$ \hspace*{\fill} \textcolor{gray}{\# extract text features and get corresponding frozen features }
                
                \vspace{0.5em}
                \STATE $\text{NN}(\tilde{z}_k^I):=\underset{q \in \mathcal{Q}_I} {\mathrm{argmin}}||\tilde{z}_k^I-q||_2$, \quad $\text{XNN}(\tilde{z}_k^I):=\tilde{z}_i^I \in \mathcal{Q}_I$, where $\tilde{z}_i^T=\text{NN}(\tilde{z}_k^T)$ \hspace*{\fill} \textcolor{gray}{\# get image NN and XNN}
                \STATE $\text{NN}(\tilde{z}_k^T):=\underset{q \in \mathcal{Q}_T}{\mathrm{argmin}}||\tilde{z}_k^T-q||_2$, \quad $\text{XNN}(\tilde{z}_k^T):=\tilde{z}_i^T \in \mathcal{Q}_T$, where $\tilde{z}_i^I=\text{NN}(\tilde{z}_k^I)$ \hspace*{\fill} \textcolor{gray}{\# get text NN and XNN }
                
                \vspace{0.5em}
                \STATE $\mathcal{L}^{\text{CLIP}}=\frac{1}{2}[\mathcal{L}_{I \rightarrow T}^{\text{CLIP}}(z_k^I, z_k^T) + \mathcal{L}_{T \rightarrow I}^{\text{CLIP}}(z_k^T, z_k^I)]$ \hspace*{\fill} \textcolor{gray}{\# CLIP loss by Eqs. (1), (2), (3) }

                \vspace{0.5em}
                \STATE $\mathcal{L}_I^{\text{NN}}= \frac{1}{2}[\mathcal{L}_{\text{NN}_I \rightarrow I}(\text{NN}(\tilde{z}_k^I), z_k^I) + \mathcal{L}_{I \rightarrow \text{NN}_I}(z_k^I, \text{NN}(\tilde{z}_k^I))]$ \hspace*{\fill} \textcolor{gray}{\# image NN supervision loss by Eqs. (4), (6)}
                \vspace{0.5em}
                \STATE $\mathcal{L}_T^{\text{NN}}= \frac{1}{2}[\mathcal{L}_{\text{NN}_T \rightarrow T}(\text{NN}(\tilde{z}_k^T), z_k^T) + \mathcal{L}_{T \rightarrow \text{NN}_T}(z_k^T, \text{NN}(\tilde{z}_k^T))]$ \hspace*{\fill} \textcolor{gray}{\# text NN supervision loss by Eqs. (5), (7)}
                \vspace{0.5em}
                \STATE $\mathcal{L}_{\text{NN}}^{\text{PING}}=\mathcal{L}_I^{\text{NN}}+\mathcal{L}_T^{\text{NN}}$ \hspace*{\fill} \textcolor{gray}{\# intra-model contrastive supervision through NN samples by Eq. (8) }

                \vspace{0.5em}
                \STATE $\mathcal{L}_I^{\text{XNN}}= \frac{1}{2}[\mathcal{L}_{\text{XNN}_I \rightarrow I} (\text{XNN}(\tilde{z}_k^I), z_k^I)+ \mathcal{L}_{I \rightarrow \text{XNN}_I}(z_k^I, \text{XNN}(\tilde{z}_k^I))]$ \hspace*{\fill} \textcolor{gray}{\# image XNN supervision loss by Eqs. (9), (11)}
                \vspace{0.5em}
                \STATE $\mathcal{L}_T^{\text{XNN}}= \frac{1}{2}[\mathcal{L}_{\text{XNN}_T \rightarrow T}(\text{XNN}(\tilde{z}_k^T), z_k^T) + \mathcal{L}_{T \rightarrow \text{XNN}_T}(z_k^T, \text{XNN}(\tilde{z}_k^T)]$ \hspace*{\fill} \textcolor{gray}{\# text XNN supervision loss by Eqs. (10), (12)}
                \vspace{0.5em}
                \STATE $\mathcal{L}_{\text{XNN}}^{\text{PING}}=\mathcal{L}_I^{\text{XNN}}+\mathcal{L}_T^{\text{XNN}}$ \hspace*{\fill} \textcolor{gray}{\# inter-model contrastive supervision through XNN samples by Eq. (13) }

                \vspace{0.5em}
                \STATE $\mathcal{L}^{\text{PING}} = (1 - \alpha) \cdot \mathcal{L}_{\text{NN}}^{\text{PING}} + \alpha \cdot \mathcal{L}_{\text{XNN}}^{\text{PING}}$ \hspace*{\fill} \textcolor{gray}{\# overall PING supervision loss by Eq. (14)}
                \vspace{0.5em}
                \STATE $\mathcal{L}^{\text{CLIP-PING}} = (1 - \lambda) \cdot \mathcal{L}^{\text{CLIP}} + \lambda \cdot \mathcal{L}^{\text{PING}}$ \hspace*{\fill} \textcolor{gray}{\# final CLIP-PING loss by Eq. (15) }

                \vspace{0.5em}
                \STATE FIFO\_UPDATE $(\mathcal{Q}_I, \tilde{z}_k^I)$ and FIFO\_UPDATE $(\mathcal{Q}_T, \tilde{z}_k^T)$
                \vspace{0.5em}
                \STATE UPDATE $(\mathcal{E}_I, \mathcal{L}^\text{CLIP-PING})$ and UPDATE $(\mathcal{E}_T, \mathcal{L}^\text{CLIP-PING})$
                \vspace{0.2em}
            \ENDFOR
        \ENDWHILE
        \RETURN lightweight vision-language model $(\mathcal{E}_I$, $\mathcal{E}_T)$
    \end{algorithmic}
    \label{alg:clip-ping}
\end{algorithm*}

%-------------------------------------------------------------------------

\section{Additional Results}\label{app:results}

\paragraph{Zero-shot classification}
In Tables~\ref{tab:zs_convnext} and \ref{tab:zs_mnv4}, we compare zero-shot image classification performance across several downstream tasks, using ConvNeXt-Pico~\cite{liu2022convnet} and MNv4-Hybrid-M~\cite{qin2024mobilenetv4} image encoders.
This comparison highlights \mbox{CLIP-PING}’s competitive performance relative to other methods, demonstrating its ability to effectively transfer knowledge across tasks with diverse encoder architectures.
Tables~\ref{tab:robustness_convnext} and \ref{tab:robustness_mnv4} shows zero-shot robustness evaluation performance of ConvNeXt-Pico~\cite{liu2022convnet} and MNv4-Hybrid-M~\cite{qin2024mobilenetv4} image encoders trained on COCO+CC3M [13, 14] dataset.

\paragraph{Linear evaluation}
In Tables~\ref{tab:linear_convnext} and \ref{tab:linear_mnv4}, we compare linear evaluation performance across several downstream tasks, using ConvNeXt-Pico~\cite{liu2022convnet} and MNv4-Hybrid-M~\cite{qin2024mobilenetv4} image encoders.
The reported values represent the average accuracy over the last 5 epochs of a total 30-epoch training process.
This comparison highlights the relative strengths of \mbox{CLIP-PING} with different encoder architectures, emphasizing its competitive performance and strong transferability across various benchmarks.

\paragraph{Impact of projection dimension}
In Table~\ref{tab:proj}, we vary the projection dimension in powers of 2, from 128 to 2048.
Our analysis reveals that a projection dimension of 256 consistently achieves the best trade-off between performance and computational efficiency across all three model pairs.
This indicates that smaller projection dimensions, such as 128, may limit the model's ability to capture rich semantic representations, while larger dimensions, such as 2048, introduce unnecessary computational overhead without substantial performance gains.
Consequently, we select 256 as the optimal projection dimension, balancing both performance and resource efficiency.

\paragraph{Random support set}
We further explore the impact of support set update strategies by comparing the default first-in-first-out (FIFO) with a random update strategy.
We find that the FIFO approach consistently outperforms random updates. Detailed results are presented in Table~\ref{tab:ss}.

\begin{table}[t]
\centering
\caption{Comparison on zero-shot classification performance. ConvNeXt-Pico~\cite{liu2022convnet} is the image encoder. The best results are marked in \textbf{bold}.}
\label{tab:zs_convnext}
\resizebox{\linewidth}{!}{
\begin{tabular}{lcccccc}
\toprule
\multicolumn{1}{c|}{Method}             & STL~\cite{coates2011analysis}    & C10~\cite{krizhevsky2009learning}      & C100~\cite{krizhevsky2009learning}      & SA-40~\cite{yao2011human} & \multicolumn{1}{c|}{IN-1K~\cite{deng2009imagenet}} & AVG  \\ \midrule
\multicolumn{7}{c}{Pre-training Dataset: COCO~\cite{lin2014microsoft} (600K)}                                                               \\ \midrule
\multicolumn{1}{l|}{CLIP}               & 69.7   & 35.5     & 9.4       & 34.6  & \multicolumn{1}{c|}{-}     & 37.3 \\
\multicolumn{1}{l|}{CLIP-D}             & \textbf{73.6}   & 40.7     & 12.0      & \textbf{41.4}  & \multicolumn{1}{c|}{-}     & \textbf{41.9} \\
\multicolumn{1}{l|}{CLIP-F}             & 70.0   & 35.4     & 10.8      & 39.0  & \multicolumn{1}{c|}{-}     & 38.8 \\
\multicolumn{1}{l|}{CLIP-KD}            & 71.1   & 35.5     & 9.2       & 36.5  & \multicolumn{1}{c|}{-}     & 38.1 \\
\multicolumn{1}{l|}{CLIP-PING (ours)}   & 71.3   & \textbf{41.6}     & \textbf{12.9}      & 41.2  & \multicolumn{1}{c|}{-}     & 41.8 \\
\multicolumn{1}{l|}{A-CLIP-PING (ours)} & \textbf{77.0}   & \textbf{57.2}     & \textbf{18.4}      & \textbf{44.0}  & \multicolumn{1}{c|}{-}     & \textbf{49.2} \\ \midrule
\multicolumn{7}{c}{Pre-training Dataset: COCO+CC3M~\cite{lin2014microsoft, sharma2018conceptual} (3M)}                                                          \\ \midrule
\multicolumn{1}{l|}{CLIP}               & 85.9   & 52.9     & 17.4      & 58.7  & \multicolumn{1}{c|}{18.6}  & 46.7 \\
\multicolumn{1}{l|}{CLIP-KD}            & 88.4   & 62.8     & 22.9      & 60.2  & \multicolumn{1}{c|}{20.7}  & 51.0 \\
\multicolumn{1}{l|}{CLIP-PING (ours)}   & \textbf{89.0}   & \textbf{63.9}     & \textbf{27.6}      & \textbf{64.0}  & \multicolumn{1}{c|}{\textbf{25.1}}  & \textbf{53.9} \\
\multicolumn{1}{l|}{A-CLIP-PING (ours)} & \textbf{91.2}   & \textbf{75.3}     & \textbf{41.6}      & \textbf{66.2}  & \multicolumn{1}{c|}{\textbf{27.1}}  & \textbf{60.3} \\ \bottomrule
\end{tabular}
}
\end{table}

\begin{table}[t]
\centering
\caption{Comparison on zero-shot robustness evaluation performance. ConvNeXt-Pico~\cite{liu2022convnet} is the image encoder. The pre-training dataset is COCO+CC3M~\cite{lin2014microsoft, sharma2018conceptual} (3M). The best results are marked in \textbf{bold}.}
\label{tab:robustness_convnext}
\resizebox{\linewidth}{!}{
\begin{tabular}{l|cccc}
\toprule
\multicolumn{1}{c|}{Method} & IN-V2~\cite{recht2019imagenet}         & IN-R~\cite{hendrycks2021many}          & IN-O~\cite{hendrycks2021natural}          & IN-S~\cite{wang2019learning}          \\ \midrule
CLIP                        & 16.1          & 17.2          & 25.8          & 7.3           \\
CLIP-KD                     & 17.6          & 19.8          & 27.1          & 9.4           \\
CLIP-PING (ours)            & \textbf{21.5} & \textbf{23.6} & \textbf{31.3} & \textbf{11.9}  \\
A-CLIP-PING (ours)          & \textbf{22.3} & \textbf{24.9} & \textbf{32.4} & \textbf{12.2} \\ \bottomrule
\end{tabular}
}
\vspace{-1.5em}
\end{table}

\begin{table}[b]
\centering
\caption{Comparison on linear evaluation performance. ConvNeXt-Pico~\cite{liu2022convnet} is the image encoder. The best results are marked in \textbf{bold}.}
\label{tab:linear_convnext}
\resizebox{\linewidth}{!}{
\begin{tabular}{lccccccccccccr}
\toprule
\multicolumn{1}{c|}{\multirow{2}{*}{Method}} & \multicolumn{4}{c|}{Mean Per Class Accuracy}                                       & \multicolumn{8}{c|}{Accuracy}                                                                                                                      & \multicolumn{1}{c}{\multirow{2}{*}{AVG}} \\ \cmidrule{2-13}
\multicolumn{1}{c|}{}                        & Pets          & Cal.          & Flow.         & \multicolumn{1}{c|}{FGVC}          & Food          & DTD           & SUN           & Cars          & STL           & C10           & C100          & \multicolumn{1}{c|}{IN-1K}         & \multicolumn{1}{c}{}                     \\ \midrule
\multicolumn{14}{c}{Pre-training Dataset: COCO~\cite{lin2014microsoft} (600K)}                                                                                                                                                                                                                                                                            \\ \midrule
\multicolumn{1}{l|}{CLIP}                    & 49.4          & 66.4          & 67.4          & \multicolumn{1}{c|}{25.5}          & 52.4          & 44.8          & 54.1          & 21.0          & 83.5          & 69.5          & 43.2          & \multicolumn{1}{c|}{-}             & 52.5                                     \\
\multicolumn{1}{l|}{CLIP-D}                  & 54.8          & 69.4          & 69.9          & \multicolumn{1}{c|}{27.3}          & 55.6          & \textbf{49.1} & \textbf{58.5} & 23.8          & \textbf{86.3} & 70.5          & 44.1          & \multicolumn{1}{c|}{-}             & 55.4                                     \\
\multicolumn{1}{l|}{CLIP-F}                  & 49.2          & 66.0          & 66.5          & \multicolumn{1}{c|}{23.4}          & 50.9          & 46.8          & 55.3          & 20.2          & 85.4          & 67.5          & 43.0          & \multicolumn{1}{c|}{-}             & 52.2                                     \\
\multicolumn{1}{l|}{CLIP-KD}                 & 52.0          & 66.7          & 66.9          & \multicolumn{1}{c|}{27.3}          & 53.1          & 44.9          & 55.2          & 23.3          & 85.4          & \textbf{71.7} & 46.5          & \multicolumn{1}{c|}{-}             & 53.9                                     \\
\multicolumn{1}{l|}{CLIP-PING}        & \textbf{58.3} & \textbf{71.3} & \textbf{72.2} & \multicolumn{1}{c|}{\textbf{30.0}} & \textbf{56.7} & 48.2          & \textbf{58.5} & \textbf{25.7} & 86.1          & 70.5          & \textbf{44.8} & \multicolumn{1}{c|}{-}             & \textbf{56.6}                            \\
\multicolumn{1}{l|}{A-CLIP-PING}      & \textbf{64.5} & \textbf{76.0} & \textbf{75.4} & \multicolumn{1}{c|}{\textbf{32.1}} & \textbf{60.5} & \textbf{50.9} & \textbf{60.3} & \textbf{28.2} & \textbf{89.8} & \textbf{77.9} & \textbf{54.2} & \multicolumn{1}{c|}{-}             & \textbf{60.9}                            \\ \midrule
\multicolumn{14}{c}{Pre-training Dataset: COCO+CC3M~\cite{lin2014microsoft, sharma2018conceptual} (3M)}                                                                                                                                                                                                                                                                         \\ \midrule
\multicolumn{1}{l|}{CLIP}                    & 63.0          & 78.7          & 79.5          & \multicolumn{1}{c|}{28.7}          & 61.0          & 59.4          & 65.3          & 26.5          & 92.5          & 73.3          & 48.9          & \multicolumn{1}{c|}{50.8}          & 60.6                                     \\
\multicolumn{1}{l|}{CLIP-KD}                 & 70.7          & 83.3          & 82.0          & \multicolumn{1}{c|}{32.2}          & 64.3          & 63.0          & 67.8          & 31.0          & 93.8          & 77.8          & 55.1          & \multicolumn{1}{c|}{56.9}          & 64.8                                     \\
\multicolumn{1}{l|}{CLIP-PING}        & \textbf{78.3} & \textbf{86.5} & \textbf{84.8} & \multicolumn{1}{c|}{\textbf{35.3}} & \textbf{66.0} & \textbf{65.0} & \textbf{69.2} & \textbf{36.8} & \textbf{94.7} & \textbf{78.2} & \textbf{55.8} & \multicolumn{1}{c|}{\textbf{59.8}} & \textbf{67.5}                            \\
\multicolumn{1}{l|}{A-CLIP-PING}      & \textbf{81.0} & \textbf{88.4} & \textbf{86.5} & \multicolumn{1}{c|}{\textbf{38.9}} & \textbf{68.9} & \textbf{65.8} & \textbf{70.2} & \textbf{39.5} & \textbf{96.1} & \textbf{87.8} & \textbf{68.7} & \multicolumn{1}{c|}{\textbf{62.5}} & \textbf{71.2}                            \\ 
\bottomrule
\end{tabular}
}
\end{table}

\paragraph{Impact of unimodal feature extractor}
In Tables~\ref{tab:coco_cc3m_vit} and \ref{tab:fex_coco}, we employ ViT-B/16~\cite{dosovitskiy2020image}, pre-trained on ImageNet~\cite{deng2009imagenet} (i.e., vit\_base\_patch16\_224) as our image teacher or feature extractor and analyze its effect on performance.
This analysis reveals that \mbox{CLIP-PING} variants consistently outperform competing methods.
Table~\ref{tab:fex_coco} summarizes results for models trained on COCO~\cite{lin2014microsoft} (600K) and Table~\ref{tab:coco_cc3m_vit} provides results for the combined dataset COCO+CC3M~\cite{lin2014microsoft, sharma2018conceptual}(3M).

\paragraph{Impact of supervision modality}
In Table~\ref{tab:supervision_source}, we investigate the impact of different supervision sources on performance using ConvNeXt-Pico~\cite{liu2022convnet} and MNv4-Hybrid-M~\cite{qin2024mobilenetv4} image encoders.
The analysis includes three setups: (1) the default configuration with both image and text supervision, (2) with only text supervision (w/o $\mathcal{L}_I^{\text{NN}}$ and $\mathcal{L}_I^{\text{XNN}}$ in Eqs.~(8) and (13)), and (3) with only image supervision (w/o $\mathcal{L}_T^{\text{NN}}$ and $\mathcal{L}_T^{\text{XNN}}$ in Eqs.~(8) and (13)).
We find that \mbox{CLIP-PING} benefits equally from both sources of supervision.

\begin{table}[t]
\centering
\caption{Comparison on zero-shot classification performance. MNv4-Hybrid-M~\cite{qin2024mobilenetv4} is the image encoder. The best results are marked in \textbf{bold}.}
\label{tab:zs_mnv4}
\resizebox{\linewidth}{!}{
\begin{tabular}{lcccccc}
\toprule
\multicolumn{1}{c|}{Method}             & STL~\cite{coates2011analysis}    & C10~\cite{krizhevsky2009learning}      & C100~\cite{krizhevsky2009learning}      & SA-40~\cite{yao2011human} & \multicolumn{1}{c|}{IN-1K~\cite{deng2009imagenet}} & AVG  \\ \midrule
\multicolumn{7}{c}{Pre-training Dataset: COCO~\cite{lin2014microsoft} (600K)}                                                               \\ \midrule
\multicolumn{1}{l|}{CLIP}               & 69.4   & 24.6     & 6.1       & 35.7  & \multicolumn{1}{c|}{-}     & 34.0 \\
\multicolumn{1}{l|}{CLIP-D}             & \textbf{73.4}   & 39.5     & 11.1      & \textbf{42.9}  & \multicolumn{1}{c|}{-}     & 41.7 \\
\multicolumn{1}{l|}{CLIP-F}             & 70.1   & 37.8     & 8.7      & 39.6  & \multicolumn{1}{c|}{-}     & 39.1 \\
\multicolumn{1}{l|}{CLIP-KD}            & 70.4   & 26.4     & 5.8       & 36.9  & \multicolumn{1}{c|}{-}     & 34.9 \\
\multicolumn{1}{l|}{CLIP-PING (ours)}   & 73.2   & \textbf{42.1}     & \textbf{12.0}      & 41.7  & \multicolumn{1}{c|}{-}     & \textbf{42.3} \\
\multicolumn{1}{l|}{A-CLIP-PING (ours)} & \textbf{79.7}   & \textbf{58.5}     & \textbf{19.8}      & \textbf{47.4}  & \multicolumn{1}{c|}{-}     & \textbf{51.4} \\ \midrule
\multicolumn{7}{c}{Pre-training Dataset: COCO+CC3M~\cite{lin2014microsoft, sharma2018conceptual} (3M)}                                                          \\ \midrule
\multicolumn{1}{l|}{CLIP}               & 85.1   & 52.3     & 20.4      & 58.1  & \multicolumn{1}{c|}{18.3}  & 46.8 \\
\multicolumn{1}{l|}{CLIP-KD}            & 88.3   & 63.4     & 24.8      & 61.1  & \multicolumn{1}{c|}{20.6}  & 51.6 \\
\multicolumn{1}{l|}{CLIP-PING (ours)}   & \textbf{89.1}   & \textbf{72.9}     & \textbf{34.1}      & \textbf{66.6}  & \multicolumn{1}{c|}{\textbf{26.2}}  & \textbf{57.8} \\
\multicolumn{1}{l|}{A-CLIP-PING (ours)} & \textbf{92.7}   & \textbf{81.0}     & \textbf{46.8}      & \textbf{68.5}  & \multicolumn{1}{c|}{\textbf{28.6}}  & \textbf{63.5} \\ \bottomrule
\end{tabular}
}
\end{table}

\begin{table}[t]
\centering
\caption{Comparison on zero-shot robustness evaluation performance. MNv4-Hybrid-M~\cite{qin2024mobilenetv4} is the image encoder. The pre-training dataset is COCO+CC3M~\cite{lin2014microsoft, sharma2018conceptual} (3M). The best results are marked in \textbf{bold}.}
\label{tab:robustness_mnv4}
\resizebox{\linewidth}{!}{
\begin{tabular}{l|cccc}
\toprule
\multicolumn{1}{c|}{Method} & IN-V2~\cite{recht2019imagenet}         & IN-R~\cite{hendrycks2021many}          & IN-O~\cite{hendrycks2021natural}          & IN-S~\cite{wang2019learning}          \\ \midrule
CLIP                        & 15.4          & 15.8          & 22.7          & 7.1           \\
CLIP-KD                     & 17.3          & 19.2          & 27.0          & 9.1           \\
CLIP-PING (ours)            & \textbf{22.2} & \textbf{23.7} & \textbf{31.4} & \textbf{11.5}  \\
A-CLIP-PING (ours)          & \textbf{24.6} & \textbf{26.3} & \textbf{33.9} & \textbf{13.8} \\ \bottomrule
\end{tabular}
}
\vspace{-1em}
\end{table}

\begin{table}[b]
\centering
\caption{Comparison on linear evaluation performance. MNv4-Hybrid-M~\cite{qin2024mobilenetv4} is the image encoder. The best results are marked in \textbf{bold}.}
\label{tab:linear_mnv4}
\resizebox{\linewidth}{!}{
\begin{tabular}{lccccccccccccr}
\toprule
\multicolumn{1}{c|}{}                         & \multicolumn{4}{c|}{Mean Per Class Accuracy}                                       & \multicolumn{8}{c|}{Accuracy}                                                                                                                                              & \multicolumn{1}{c}{\multirow{2}{*}{AVG}} \\ \cmidrule{2-13}
\multicolumn{1}{c|}{\multirow{-2}{*}{Method}} & Pets          & Cal.          & Flow.         & \multicolumn{1}{c|}{FGVC}          & Food          & DTD           & SUN           & Cars          & STL           & C10           & C100          & \multicolumn{1}{c|}{IN-1K}         & \multicolumn{1}{c}{}                     \\ \midrule
\multicolumn{14}{c}{Pre-training Dataset: COCO~\cite{lin2014microsoft} (600K)}                                                                                                                                                                                                                                                                                                      \\ \midrule
\multicolumn{1}{l|}{CLIP}                     & 37.4          & 57.8          & 56.2          & \multicolumn{1}{c|}{14.7}          & 36.7                                  & 42.8          & 47.4          & 12.7          & 80.4          & 49.0          & 20.4          & \multicolumn{1}{c|}{-}             & 41.4                                      \\
\multicolumn{1}{l|}{CLIP-D}                   & 41.3          & \textbf{62.0} & 57.0          & \multicolumn{1}{c|}{16.3}          & 38.8                                  & \textbf{46.6} & 51.2          & \textbf{14.8} & 82.9          & 53.1          & 24.8          & \multicolumn{1}{c|}{-}             & 44.4                                      \\
\multicolumn{1}{l|}{CLIP-F}                   & 35.7          & 58.0          & 55.5          & \multicolumn{1}{c|}{15.7}          & 35.4                                  & 44.6          & 47.7          & 12.5          & 80.3          & 50.0          & 21.7          & \multicolumn{1}{c|}{-}             & 41.6                                      \\
\multicolumn{1}{l|}{CLIP-KD}                  & 42.7          & 63.0          & 63.1          & \multicolumn{1}{c|}{18.4}          & 41.6                                  & 46.9          & 52.0          & 16.2          & 82.5          & 53.2          & 24.5          & \multicolumn{1}{c|}{-}             & 45.8                                      \\
\multicolumn{1}{l|}{CLIP-PING}         & \textbf{49.9} & 66.7          & \textbf{70.0} & \multicolumn{1}{c|}{\textbf{21.6}} & \textbf{48.1}                         & 48.3          & \textbf{54.4} & 19.4          & \textbf{84.1} & \textbf{55.4} & \textbf{27.9} & \multicolumn{1}{c|}{-}             & \textbf{49.6}                             \\
\multicolumn{1}{l|}{A-CLIP-PING}       & \textbf{57.2} & \textbf{72.8} & \textbf{73.2} & \multicolumn{1}{c|}{\textbf{24.5}} & \textbf{51.8}                         & \textbf{51.5} & \textbf{58.1} & \textbf{23.5} & \textbf{85.9} & \textbf{66.7} & \textbf{40.6} & \multicolumn{1}{c|}{-}             & \textbf{55.1}                             \\ \midrule
\multicolumn{14}{c}{Pre-training Dataset: COCO+CC3M~\cite{lin2014microsoft, sharma2018conceptual} (3M)}                                                                                                                                                                                                                                                                                                   \\ \midrule
\multicolumn{1}{l|}{CLIP}                     & 49.0          & 75.8          & 72.6          & \multicolumn{1}{c|}{21.3}          & 46.7                                  & 55.3          & 61.3          & 19.4          & 79.9          & 50.3          & 23.5          & \multicolumn{1}{c|}{49.6}          & 50.4                                      \\
\multicolumn{1}{l|}{CLIP-KD}                  & 55.8          & 79.2          & 76.3          & \multicolumn{1}{c|}{24.9}          & 51.4                                  & 57.7          & 63.6          & 22.4          & 86.2          & 58.8          & 32.5          & \multicolumn{1}{c|}{54.5}          & 55.3                                      \\
\multicolumn{1}{l|}{CLIP-PING}         & \textbf{66.8} & \textbf{83.0} & \textbf{81.8} & \multicolumn{1}{c|}{\textbf{29.9}} & \textbf{57.5} & \textbf{59.7} & \textbf{66.3} & \textbf{30.5} & \textbf{90.6} & \textbf{58.1} & \textbf{36.4} & \multicolumn{1}{c|}{\textbf{61.8}} & \textbf{60.2}                             \\
\multicolumn{1}{l|}{A-CLIP-PING}       & \textbf{70.3} & \textbf{85.2} & \textbf{81.6} & \multicolumn{1}{c|}{\textbf{31.7}} & \textbf{58.5} & \textbf{60.0} & \textbf{67.4} & \textbf{32.7} & \textbf{92.9} & \textbf{69.2} & \textbf{43.4} & \multicolumn{1}{c|}{\textbf{64.4}} & \textbf{63.1}                             \\ \bottomrule
\end{tabular}
}
\end{table}

\begin{table*}
\centering
\caption{Impact of projection dimension ($d$) on retrieval performance for Flickr30K~\cite{young2014image} and average zero-shot top-1 classification accuracy across 4 datasets, using the image encoders pre-trained on COCO~\cite{lin2014microsoft} dataset.}
\label{tab:proj}
\begin{subtable}[t]{0.3\textwidth}
\centering
\caption{\footnotesize With ViT-XS~\cite{dosovitskiy2020image} image encoder.}
\label{tab:proj_dim_vit}
\resizebox{\linewidth}{!}{
\begin{tabular}{c|cc|c}
\toprule
\multirow{2}{*}{$d$} & \multicolumn{2}{c|}{Flickr30K~\cite{young2014image}} & ZS   \\
                       & I2T@1          & T2I@1         & AVG  \\ \midrule
128                    & 26.3           & 19.1          & 39.7 \\
256                    & 28.1           & 20.2          & \textbf{41.2} \\
512                    & \textbf{28.5}           & \textbf{20.8}          & 39.2 \\
1024                    & 25.3           & 19.5          & 40.8 \\
2048                    & 25.9           & 20.2          & 39.4 \\ \bottomrule
\end{tabular}
}
\end{subtable}
\hspace{\fill}
\begin{subtable}[t]{0.3\textwidth}
\centering
\caption{\footnotesize With ConvNeXt-Pico~\cite{liu2022convnet} image encoder.}
\label{tab:proj_dim_convnext}
\resizebox{\linewidth}{!}{
\begin{tabular}{c|cc|c}
\toprule
\multirow{2}{*}{$d$} & \multicolumn{2}{c|}{Flickr30K~\cite{young2014image}} & ZS   \\
                       & I2T@1          & T2I@1         & AVG  \\ \midrule
128                    & 28.1           & 20.7          & 40.6 \\
256                    & \textbf{29.6}  & 20.8          & 41.8 \\
512                    & 28.2           & 20.8          & 41.8 \\
1024                   & 27.2           & \textbf{22.2} & 43.2 \\
2048                   & 27.6           & 21.3          & \textbf{44.1} \\ \bottomrule
\end{tabular}
}
\end{subtable}
\hspace{\fill}
\begin{subtable}[t]{0.3\textwidth}
\centering
\caption{\footnotesize With MNv4-Hybrid-M~\cite{qin2024mobilenetv4} image encoder.}
\label{tab:proj_dim_mnv4}
\resizebox{\linewidth}{!}{
\begin{tabular}{c|cc|c}
\toprule
\multirow{2}{*}{$d$} & \multicolumn{2}{c|}{Flickr30K~\cite{young2014image}} & ZS   \\
                       & I2T@1          & T2I@1         & AVG  \\ \midrule
128                    & 27.4           & 21.0          & 40.3 \\
256                    & \textbf{29.9}  & 21.5          & \textbf{42.3} \\
512                    & 28.4           & 20.9          & 42.2 \\
1024                   & 28.8           & 21.1          & 41.5 \\
2048                   & 28.7           & \textbf{21.6} & 41.9 \\ \bottomrule
\end{tabular}
}
\end{subtable}
\end{table*}

\begin{table*}
\centering
\caption{Impact of support set update strategy on retrieval performance for Flickr30K~\cite{young2014image} and average zero-shot top-1 classification accuracy across 4 datasets, using the image encoders pre-trained on COCO~\cite{lin2014microsoft} dataset.}
\label{tab:ss}
\begin{subtable}[t]{0.3\textwidth}
\centering
\caption{\footnotesize With ViT-XS~\cite{dosovitskiy2020image} image encoder.}
\label{tab:ss_vit}
\resizebox{\linewidth}{!}{
\begin{tabular}{c|cc|c}
\toprule
\multirow{2}{*}{\begin{tabular}[c]{@{}c@{}}Support Set\\ Update\end{tabular}} & \multicolumn{2}{c|}{Flickr30K~\cite{young2014image}} & ZS   \\
                             & I2T@1          & T2I@1         & AVG  \\ \midrule
FIFO                         & \textbf{28.1}           & \textbf{20.2}          & \textbf{41.2} \\
Random                     & 27.1           & 19.1          & 39.6 \\ \bottomrule
\end{tabular}
}
\end{subtable}
\hspace{\fill}
\begin{subtable}[t]{0.3\textwidth}
\centering
\caption{\footnotesize With ConvNeXt-Pico~\cite{liu2022convnet} image encoder.}
\label{tab:ss_convnext}
\resizebox{\linewidth}{!}{
\begin{tabular}{c|cc|c}
\toprule
\multirow{2}{*}{\begin{tabular}[c]{@{}c@{}}Support Set\\ Update\end{tabular}} & \multicolumn{2}{c|}{Flickr30K~\cite{young2014image}} & ZS   \\
                             & I2T@1          & T2I@1         & AVG  \\ \midrule
FIFO                         & \textbf{29.6}           & \textbf{20.8}          & \textbf{41.8} \\
Random                      & 26.6           & 20.9          & 41.4 \\ \bottomrule
\end{tabular}
}
\end{subtable}
\hspace{\fill}
\begin{subtable}[t]{0.3\textwidth}
\centering
\caption{\footnotesize With MNv4-Hybrid-M~\cite{qin2024mobilenetv4} image encoder.}
\label{tab:ss_mnv4}
\resizebox{\linewidth}{!}{
\begin{tabular}{c|cc|c}
\toprule
\multirow{2}{*}{\begin{tabular}[c]{@{}c@{}}Support Set\\ Update\end{tabular}} & \multicolumn{2}{c|}{Flickr30K~\cite{young2014image}} & ZS   \\
                             & I2T@1          & T2I@1         & AVG  \\ \midrule
FIFO                         & \textbf{29.9}           & \textbf{21.5}          & \textbf{42.3} \\
Random                      & 25.7           & 19.8          & 40.4 \\ \bottomrule
\end{tabular}
}
\end{subtable}
\end{table*}

\begin{table*}[t]
\centering
\caption{Effect of ViT-B/16~\cite{dosovitskiy2020image} on retrieval performance for Flickr30K~\cite{young2014image} and average zero-shot top-1 classification accuracy across 5 datasets, using image encoders pre-trained on COCO+CC3M~\cite{lin2014microsoft, sharma2018conceptual} dataset. The best results are marked in \textbf{bold}. Memory usage is recorded on a single NVIDIA A100 GPU.}
\label{tab:coco_cc3m_vit}
\begin{subtable}[t]{0.32\textwidth}
\centering
\caption{\footnotesize With ViT-XS~\cite{dosovitskiy2020image} image encoder.}
\label{tab:vit_vit}
\resizebox{\linewidth}{!}{
\begin{tabular}{l|c|cc|c}
\toprule
\multicolumn{1}{c|}{\multirow{2}{*}{Method}} & \multirow{2}{*}{\begin{tabular}[c]{@{}c@{}}Memory\\ (MiB)$\downarrow$\end{tabular}} & \multicolumn{2}{c|}{Flickr30K~\cite{young2014image}} & \multirow{2}{*}{\begin{tabular}[c]{@{}c@{}}ZS\\ AVG\end{tabular}} \\
\multicolumn{1}{c|}{}                        &                                                                         & I2T@1          & T2I@1         &                                                                   \\ \midrule
CLIP-KD                                      & 17087                                                                   & 46.1           & 34.1          & 52.7                                                               \\
CLIP-PING                                    & 11689                                                                   & \textbf{46.5}           & \textbf{34.5}          & \textbf{54.1}                                                               \\
A-CLIP-PING                                  & 17273                                                                   & \textbf{48.6}           & \textbf{35.8}          & \textbf{55.6}                                                               \\ \bottomrule
\end{tabular}
}
\end{subtable}
\hspace{\fill}
\begin{subtable}[t]{0.32\textwidth}
\centering
\caption{\footnotesize With ConvNeXt-Pico~\cite{liu2022convnet} image encoder.}
\label{tab:vit_convnext}
\resizebox{\linewidth}{!}{
\begin{tabular}{l|c|cc|c}
\toprule
\multicolumn{1}{c|}{\multirow{2}{*}{Method}} & \multirow{2}{*}{\begin{tabular}[c]{@{}c@{}}Memory\\ (MiB)$\downarrow$\end{tabular}} & \multicolumn{2}{c|}{Flickr30K~\cite{young2014image}} & \multirow{2}{*}{\begin{tabular}[c]{@{}c@{}}ZS\\ AVG\end{tabular}} \\
\multicolumn{1}{c|}{}                        &                                                                         & I2T@1          & T2I@1         &                                                                   \\ \midrule
CLIP-KD                                      & 21375                                                                   & \textbf{53.1}           & \textbf{39.4}          & \textbf{54.9}                                                               \\
CLIP-PING                                    & 18459                                                                   & 52.7           & 38.1          & \textbf{54.9}                                                               \\
A-CLIP-PING                                  & 21563                                                                   & \textbf{54.6}           & \textbf{39.6}          & \textbf{57.3}                                                               \\ \bottomrule
\end{tabular}
}
\end{subtable}
\hspace{\fill}
\begin{subtable}[t]{0.32\textwidth}
\centering
\caption{\footnotesize With MNv4-Hybrid-M~\cite{qin2024mobilenetv4} image encoder.}
\label{tab:vit_mnv4}
\resizebox{\linewidth}{!}{
\begin{tabular}{l|c|cc|c}
\toprule
\multicolumn{1}{c|}{\multirow{2}{*}{Method}} & \multirow{2}{*}{\begin{tabular}[c]{@{}c@{}}Memory\\ (MiB)$\downarrow$\end{tabular}} & \multicolumn{2}{c|}{Flickr30K~\cite{young2014image}} & \multirow{2}{*}{\begin{tabular}[c]{@{}c@{}}ZS\\ AVG\end{tabular}} \\
\multicolumn{1}{c|}{}                        &                                                                         & I2T@1          & T2I@1         &                                                                   \\ \midrule
CLIP-KD                                      & 21609                                                                   & \textbf{52.2}           & 40.1          & 55.7                                                               \\
CLIP-PING                                    & 18299                                                                   & 51.4           & \textbf{40.7}          & \textbf{56.2}                                                               \\
A-CLIP-PING                                  & 22371                                                                   & \textbf{53.8}           & \textbf{41.8}          & \textbf{60.0}                                                               \\ \bottomrule
\end{tabular}
}
\end{subtable}
\vspace{-1em}
\end{table*}

\begin{table}[t]
\centering
\caption{Effect of ViT-B/16~\cite{dosovitskiy2020image} on retrieval performance for Flickr30K~\cite{young2014image} and average zero-shot top-1 classification accuracy across 4 datasets, using image encoders pre-trained on COCO~\cite{lin2014microsoft} dataset. The best results are marked in \textbf{bold}.}
\label{tab:fex_coco}
\begin{subtable}[t]{0.23\textwidth}
\centering
\caption{\footnotesize  With ConvNeXt-Pico~\cite{liu2022convnet}.}
\label{tab:fex_convnext}
\resizebox{\linewidth}{!}{
\begin{tabular}{l|cc|c}
\toprule
\multicolumn{1}{c|}{\multirow{2}{*}{Method}} & \multicolumn{2}{c|}{Flickr30K~\cite{young2014image}} & ZS   \\
\multicolumn{1}{c|}{}                        & I2T@1          & T2I@1         & AVG  \\ \midrule
CLIP-D                                       & 25.1           & 19.1          & \textbf{42.6}  \\
CLIP-F                                       & 24.9           & 17.4          & 41.6  \\
CLIP-KD                                      & \textbf{28.0}           & \textbf{21.2}          & 40.4 \\
CLIP-PING                                    & 27.9           & 20.0          & 41.5 \\
A-CLIP-PING                                  & \textbf{29.4}           & \textbf{21.9}          & \textbf{45.6} \\ \bottomrule
\end{tabular}
}
\end{subtable}
\hspace{\fill}
\begin{subtable}[t]{0.23\textwidth}
\centering
\caption{\footnotesize With MNv4-Hybrid-M~\cite{qin2024mobilenetv4}.}
\label{tab:fex_mnv4}
\resizebox{\linewidth}{!}{
\begin{tabular}{l|cc|c}
\toprule
\multicolumn{1}{c|}{\multirow{2}{*}{Method}} & \multicolumn{2}{c|}{Flickr30K~\cite{young2014image}} & ZS   \\
\multicolumn{1}{c|}{}                        & I2T@1          & T2I@1         & AVG  \\ \midrule
CLIP-D                                       & 22.6           & 17.4          & 38.2 \\
CLIP-F                                       & 19.6           & 14.2          & 33.4  \\
CLIP-KD                                      & 27.2           & \textbf{21.0}          & 39.0 \\
CLIP-PING                                    & \textbf{27.3}           & 20.5          & \textbf{39.3} \\
A-CLIP-PING                                  & \textbf{30.8}           & \textbf{22.5}          & \textbf{42.8} \\ \bottomrule
\end{tabular}
}
\end{subtable}
\end{table}

\begin{table}[t]
\centering
\caption{Ablation on supervision sources for retrieval performance on Flickr30K~\cite{young2014image} and average zero-shot top-1 classification accuracy across 4 datasets, using image encoders pre-trained on COCO~\cite{lin2014microsoft} dataset.}
\label{tab:supervision_source}
\begin{subtable}[t]{0.23\textwidth}
\centering
\caption{\footnotesize  With ConvNeXt-Pico~\cite{liu2022convnet}.}
\label{tab:nn_convnext}
\resizebox{\linewidth}{!}{
\begin{tabular}{c|cc|c}
\toprule
\multirow{2}{*}{\begin{tabular}[c]{@{}c@{}}Super-\\ vision\end{tabular}} & \multicolumn{2}{c|}{Flickr30K~\cite{young2014image}} & ZS   \\
                             & I2T@1          & T2I@1         & AVG  \\ \midrule
both                         & \textbf{29.6}           & \textbf{20.8}          & \textbf{41.8} \\
txt only                      & 19.1           & 15.8          & 40.5 \\
img only                      & 25.6           & 18.4          & 42.2 \\ \bottomrule
\end{tabular}
}
\end{subtable}
\hspace{\fill}
\begin{subtable}[t]{0.23\textwidth}
\centering
\caption{\footnotesize With MNv4-Hybrid-M~\cite{qin2024mobilenetv4}.}
\label{tab:nn_mnv4}
\resizebox{\linewidth}{!}{
\begin{tabular}{c|cc|c}
\toprule
\multirow{2}{*}{\begin{tabular}[c]{@{}c@{}}Super-\\ vision\end{tabular}} & \multicolumn{2}{c|}{Flickr30K~\cite{young2014image}} & ZS   \\
                             & I2T@1          & T2I@1         & AVG  \\ \midrule
both                         & \textbf{29.9}           & \textbf{21.5}          & \textbf{42.3} \\
txt only                      & 21.2           & 17.2          & 37.5 \\
img only                      & 25.4           & 18.4          & 39.4 \\ \bottomrule
\end{tabular}
}
\end{subtable}
\vspace{-1.3em}
\end{table}

%-------------------------------------------------------------------------

\bibliographystyle{IEEEtran}
\bibliography{main}

\end{document}